\documentclass[lettersize,journal]{IEEEtran}
\pdfoutput=1
\usepackage{amsmath,amsfonts}
\usepackage{algorithmic}
\usepackage{algorithm}
\usepackage{array}
\usepackage{subfig}
\usepackage{caption}
\usepackage{textcomp}
\usepackage{stfloats}
\usepackage{url}
\usepackage{verbatim}
\usepackage{graphicx}
\usepackage{cite}
\usepackage{epsfig}
\usepackage{booktabs}
\usepackage[dvipsnames, table]{xcolor}
\usepackage{multirow}
\usepackage{multicol}
\usepackage{ragged2e}
\usepackage{arydshln}
\usepackage{indentfirst}
\usepackage{footmisc}
\usepackage[normalem]{ulem}
\usepackage{moresize}
\usepackage{amsthm}
\usepackage[accsupp]{axessibility}
\usepackage{hyperref}
\hypersetup{
    colorlinks=true,
    linkcolor=RubineRed,
    filecolor=magenta,      
    urlcolor=RubineRed,
    pdftitle={Overleaf Example},
    pdfpagemode=FullScreen,
}
\DeclareMathOperator*{\argmax}{argmax}

\definecolor{LAA-Net}{HTML}{ffadad}
\definecolor{MultiAtt.}{HTML}{a0c4ff}
\definecolor{SBI}{HTML}{bdb2ff}
\definecolor{Xception}{HTML}{caffbf}
\definecolor{RECCE}{HTML}{ffd6a5}
\definecolor{CADDM}{HTML}{ffc6ff}
\definecolor{FAViT}{HTML}{94d2bd}
\definecolor{ForensicsAdapter}{HTML}{8eecf5}
\definecolor{FakeFormer}{HTML}{684756}
\definecolor{FakeSwin}{HTML}{96705b}
\definecolor{ViT}{HTML}{9A8C98}
\definecolor{Swin}{HTML}{F2E9E4}
\definecolor{LAA-Net_SBI}{HTML}{FFADAD}
\definecolor{CADDM_FakeFormer}{HTML}{FFD6A5}
\definecolor{EFN_SBI}{HTML}{bdb2ff}
\definecolor{Xception_SBI}{HTML}{fcac5d}

\definecolor{darkgreen}{rgb}{0,0.5,0} 
\definecolor{purple}{rgb}{1,0,1} 

\newcolumntype{H}{>{\setbox0=\hbox\bgroup}c<{\egroup}@{}}

\begin{document}

\title{LAA-X: Unified Localized Artifact Attention for Quality-Agnostic and Generalizable Face Forgery Detection}

\author{Dat NGUYEN$^{\ast}$,  Enjie GHORBEL, Anis KACEM,~\IEEEmembership{Member,~IEEE}, Marcella ASTRID$^\dagger$,~\IEEEmembership{Member,~IEEE}, Djamila AOUADA,~\IEEEmembership{Senior Member,~IEEE}
\thanks{\textit{($^{\ast}$ Corresponding author: Dat NGUYEN)}}
\thanks{Dat NGUYEN, Anis KACEM, and Djamila AOUADA are with the CVI$^2$, SnT, University of Luxembourg, Luxembourg (emails: dat.nguyen@uni.lu; anis.kacem@uni.lu; djamila.aouada@uni.lu).}
\thanks{Enjie GHORBEL has a double affiliation. She is with CRISTAL laboratory, ENSI, University of Manouba, Tunisia  and SnT, University of Luxembourg, Luxembourg (email: enjie.ghorbel@ensi-uma.tn).}
\thanks{$\dagger$ This work was done while Marcella ASTRID was a Postdoctoral Researcher at CVI$^2$, SnT, University of Luxembourg, Luxembourg. (email: marcella.astrid@gmail.com)}
}

\markboth{Journal of \LaTeX\ Class Files,~Vol.~14, No.~8, August~2021}%
{Shell \MakeLowercase{\textit{et al.}}: A Sample Article Using IEEEtran.cls for IEEE Journals}


\maketitle
\makeatletter
\newcommand*\bigcdot{\mathpalette\bigcdot@{1.2}}
\newcommand*\bigcdot@[2]{\mathbin{\vcenter{\hbox{\scalebox{#2}{$\m@th#1\bullet$}}}}}
\makeatother

\newcommand{\kibitz}[2]{\ifnum\Comments=1\textcolor{#1}{#2}\fi}

\definecolor{darkgreen}{rgb}{0,0.5,0} 
\definecolor{purple}{rgb}{1,0,1} 
\newcommand{\EG}[1]{\kibitz{red}{Enjie: #1}}
\newcommand{\DN}[1]{\kibitz{darkgreen}{Dat: #1}}
\newcommand{\IS}[1]{\kibitz{pink}{Inder: #1}}
\newcommand{\NM}[1]{\kibitz{purple}{Nesryne: #1}}
\newcommand{\DA}[1]{\kibitz{blue}{[DA: #1}]}
\newcommand{\AK}[1]{\kibitz{cyan}{Anis: #1}}

\newtheorem{Definition}{Definition}
\newtheorem{SubDefinition}{Definition}[Definition]

\begin{abstract}
In this paper, we propose Localized Artifact Attention X (LAA-X), a novel deepfake detection framework that is both robust to high-quality forgeries and capable of generalizing to unseen manipulations. Existing approaches typically rely on binary classifiers coupled with implicit attention mechanisms, which often fail to generalize beyond known manipulations. In contrast, LAA-X introduces an explicit attention strategy based on a multi-task learning framework combined with blending-based data synthesis. Auxiliary tasks are designed to guide the model toward localized, artifact-prone (i.e., vulnerable) regions. The proposed framework is compatible with both CNN and transformer backbones, resulting in two different versions, namely, LAA-Net and LAA-Former, respectively. Despite being trained only on real and pseudo-fake samples, LAA-X competes with state-of-the-art methods across multiple benchmarks. Code and pre-trained weights for LAA-Net\footnote{\url{https://github.com/10Ring/LAA-Net}} and LAA-Former\footnote{\url{https://github.com/10Ring/LAA-Former}} are publicly available.
\end{abstract}


\section{Introduction}
\label{sec:intro}

Advances in generative modeling have significantly simplified the automated creation of photorealistic facial forgeries, commonly known as deepfakes.  While this technology supports creative and educational applications, its misuse poses serious political and societal threats~\cite{russia-ukraine-war, african-leaders}. Unfortunately, detecting forged images with the naked eye is becoming extremely challenging, particularly when encountering the most realistic ones, often referred to as High-Quality (HQ) deepfakes.  As a result, there is a pressing need to introduce methods that can automatically spot deepfakes, including HQ samples. 

Earlier deepfake detectors~\cite{ff++, mesonet, fake_spotter, f3-net, capsulenet, HCiT, efn_vit, fdftnet, conv_pooling_transformer, ViT_dropout} typically make use of Deep Neural Networks (DNNs) under a binary classification setting. Despite being promising, these approaches exhibit two fundamental weaknesses, namely:

\begin{figure}[!t]
    \centering
    \subfloat[]{
        \includegraphics[width=0.73\linewidth]{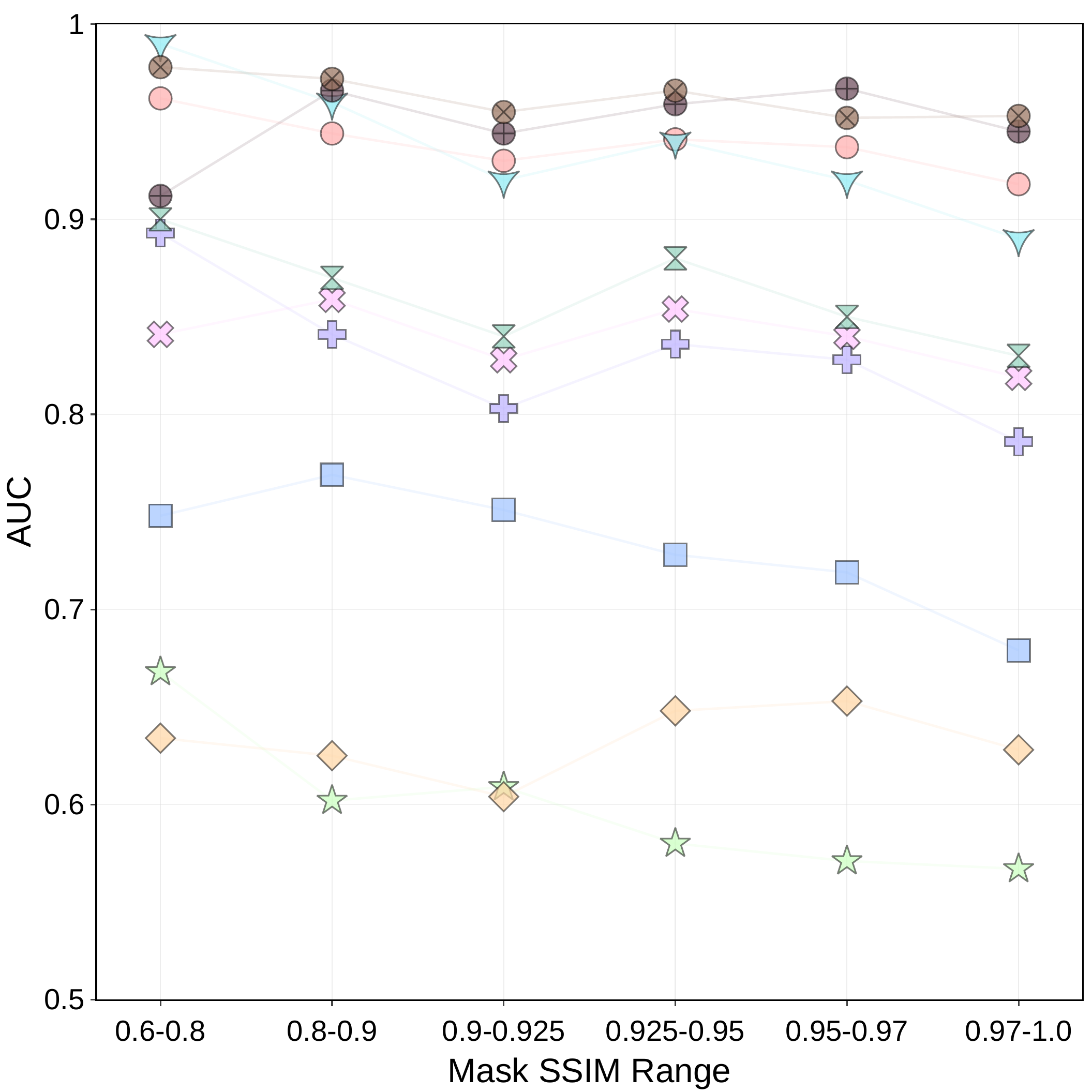}
    }
    \vspace{-4mm}
    \subfloat[]{
        \includegraphics[width=0.73\linewidth]{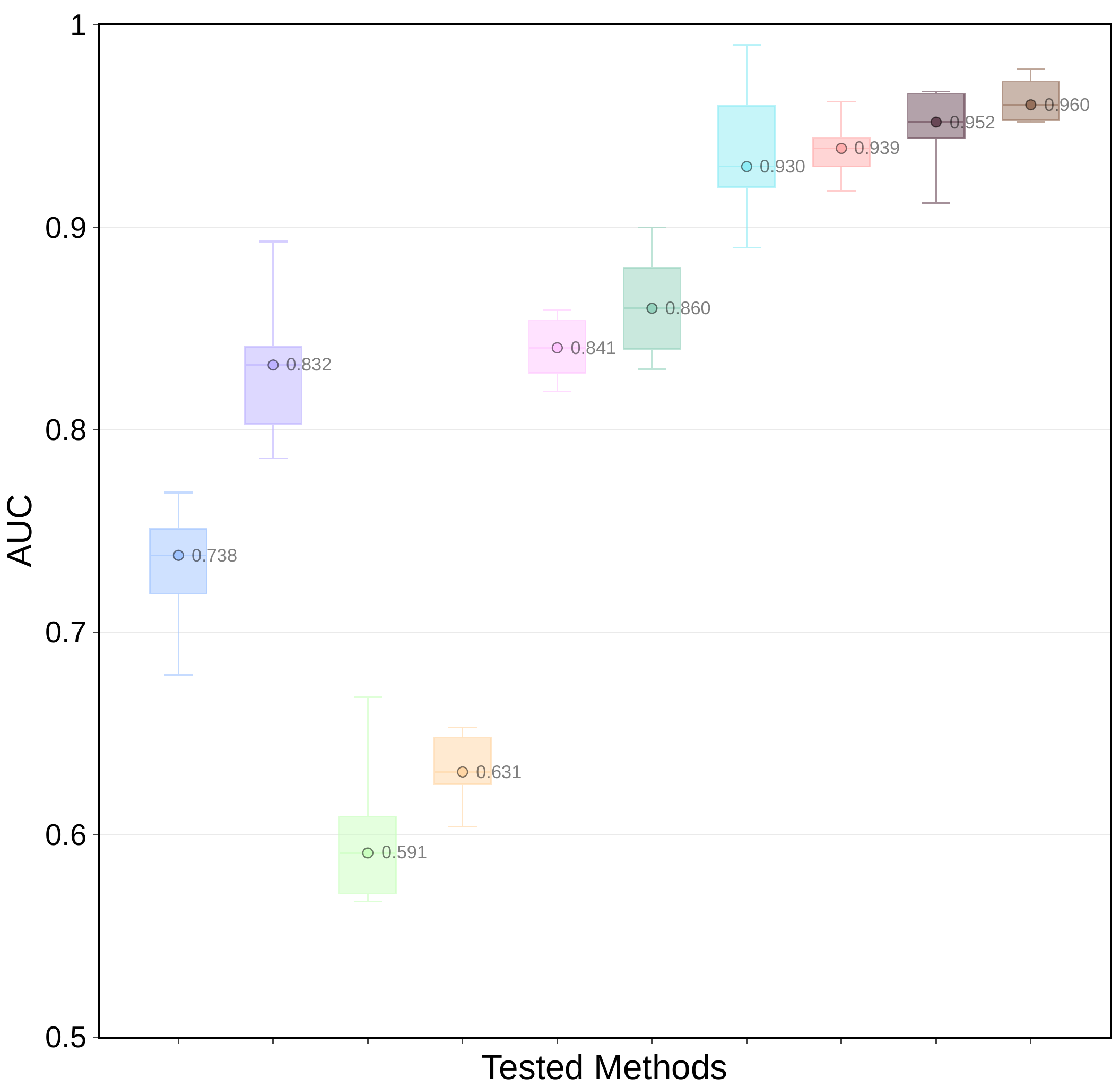}
    }
    \vspace{-1mm}
    \caption{Comparison of LAA-Net(\textcolor{LAA-Net}{$\bigcdot$}), LAA-Former(\textcolor{FakeFormer}{$\bigcdot$}), and LAA-Swin(\textcolor{FakeSwin}{$\bigcdot$}) with respect to existing methods, namely, Multi-attentional(\textcolor{MultiAtt.}{$\bigcdot$})~\cite{multi-attentional}, SBI(\textcolor{SBI}{$\bigcdot$})~\cite{sbi}, Xception(\textcolor{Xception}{$\bigcdot$})~\cite{ff++}, RECCE(\textcolor{RECCE}{$\bigcdot$})~\cite{ete_recons}, CADDM(\textcolor{CADDM}{$\bigcdot$})~\cite{caddm}, FAViT(\textcolor{FAViT}{$\bigcdot$})~\cite{FAViT}, ForensicsAdapter(\textcolor{ForensicsAdapter}{$\bigcdot$})~\cite{forensicsadapter} using (a) the AUC performance with respect to different ranges of Mask-SSIM, and (b) its associated boxplots.  *The results were obtained using the official source codes pretrained on FF+~\cite{ff++} and testing on Celeb-DFv2~\cite{celeb_df}. Figure best viewed in colors.}
    \vspace{-3mm}
    \label{fig:motivation}
\end{figure}

\textbf{(1) Limited generalization -} As highlighted in numerous studies~\cite{sladd, fxray, caddm, sbi, cstency_learning, laa_net, forensicsadapter, fakestormer}, standard deep binary classifiers tend to overfit the manipulation-specific traces present in the training set, failing to generalize to unseen forgery artifacts.

\textbf{(2) Limited robustness to HQ deepfakes -} Most previous works employ vanilla CNN and/or ViT backbones for feature extraction. CNN-based architectures such as EfficientNet~\cite{efn_net}, XceptionNet~\cite{xception}, and ResNet~\cite{resnet} are known to progressively dilute local information through successive convolutions~\cite{multi-attentional, laa_net, sfdg}. ViTs~\cite{ViT, deit}, while effective at modeling long-range dependencies, lack dedicated mechanisms to capture fine-grained, spatially localized features~\cite{twin, swin, nat, attentionflow}.
As a result, these methods show poor robustness to HQ deepfakes, which are typically characterized by subtle artifacts.

Recently, tremendous efforts have been made to mitigate generalization issues~\cite{ete_recons, ucf,fxray, sbi, sladd, cstency_learning, ost, fakestormer}. Diverse approaches have been proposed in the literature, such as multi-task learning~\cite{ete_recons, ucf}, data synthesis~\cite{fxray, sbi, sladd, cstency_learning, ost, fakestormer}, or adaptation from large pre-trained models~\cite{GiantShoulder, forensicsadapter, UDD}. Although promising, these methods usually rely on the aforementioned standard backbones, which tend to overlook local representations, resulting in limited performance when dealing with HQ forgeries.

On the other hand, some attempts have been made to improve the robustness to HQ deepfakes~\cite{multi-attentional, sfdg, dfdt} by imposing the extraction of local features through appropriate attention mechanisms.  However, these mechanisms are \textit{implicit}, with no guarantee of modeling genuinely localized yet artifact-relevant features. Moreover, these models typically rely on standard binary classifiers trained with real and deepfake images, showing inevitably degraded generalization capabilities to unseen manipulations.

Hence, our goal is to address the detection of high-quality deepfakes and, at the same time, improve the generalization performance. We posit that both objectives can be achieved by introducing an \textit{explicit} fine-grained attention mechanism within a multi-task learning framework, supervised by appropriate pseudo-labels to support generalization. 
Herein, this paper introduces an explicit fine-grained framework for deepfake detection, called Localized Artifact Attention X (LAA-X), that is generic to unseen manipulations yet robust to HQ deepfakes. LAA-X proposes a multi-task learning framework with auxiliary tasks that enable focusing on vulnerable regions\footnote{A more formal definition is given in Section~\ref{sec:vulnerability}}. By vulnerable regions, we mean small image portions that are most likely to carry blending artifacts resulting from face manipulations. The proposed framework is compatible with both CNN and transformer backbones, showing improved performance on several well-known benchmarks.
This paper is an extended version of our previous work, termed LAA-Net~\cite{laa_net}, where a CNN-based multi-task learning framework for deepfake detection has been proposed.  
The initial version of LAA-Net, proposed in~\cite{laa_net}, primarily models local dependencies as it relies on a CNN architecture. As a result, it has limited capabilities for reasoning over spatially distant regions, which are often interrelated in facial images. To address this, we generalize our framework to the transformer architecture, resulting in \textit{LAA-Former}, which combines global context modeling with explicit local attention. 
Transformers excel at capturing global dependencies~\cite{nat, swin, ViT, twin, attentionflow} but often overlook fine-grained artifacts due to their broad receptive fields and patch-level abstraction.
To overcome this, we introduce a lightweight and plug-and-play \textit{Learning-based Local Attention (L2-Att)} module that generalizes the vulnerability concept from pixels to patches, enabling transformers to explicitly attend to vulnerable areas while preserving their ability to capture long-range relationships. This integration unifies explicit local supervision and global relational reasoning within a single framework. As such, \textit{LAA-Net} and \textit{LAA-Former} together form two different versions of the proposed \textit{LAA-X} framework, where $X$ refers to the nature of the architecture.
As reflected in Figure~\ref{fig:motivation}, LAA-X achieves better and more stable AUC performance on Celeb-DF dataset~\cite{celeb_df}  with respect to existing methods~\cite{multi-attentional, sbi, ff++, ete_recons, caddm, FAViT, forensicsadapter}, especially when facing high-quality deepfakes. Specifically, we quantify the quality of deepfakes using the well-known Mask Structural SIMilarity (Mask-SSIM\footnote{The Mask-SSIM~\cite{mssim_pose, celeb_df} is computed by calculating the similarity in the head region between the fake image and its original version using the SSIM score introduced in~\cite{ssim_1}. Hence, a higher Mask-SSIM score corresponds to a deepfake of higher quality. \label{mssim_defi}}). Additional experiments on several benchmarks~\cite{celeb_df, dfd, wdf, dfdcp, dfdc, df40, diffusionface} also demonstrate that LAA-X outperforms the state-of-the-art (SOTA).

In summary,  the contributions of this extended version as compared to~\cite{laa_net} are the following:

\begin{enumerate}
    \item A unified deepfake detection framework (LAA-X) that is generic yet robust to HQ facial forgeries. LAA-X is compatible with both CNN and Transformer architectures and is trained using real data only.
    \item An extension of the proposed explicit attention to transformer backbones termed L2-Att that generalizes vulnerability-driven modeling from pixels to patches, enabling complementary local-global reasoning.
    \item Deeper and more extensive experiments showing consistent state-of-the-art performance and robustness on eight challenging benchmarks, namely FF++~\cite{ff++}, CDF2~\cite{celeb_df}, DFD~\cite{dfd}, DFW~\cite{wdf}, DFDCP~\cite{dfdcp}, DFDC~\cite{dfdc}, DF40~\cite{df40}, and DiffSwap~\cite{diffusionface} for both LAA-Net and LAA-Fomer.
\end{enumerate}

\vspace{1mm}
\noindent\textbf{Paper Organization.}
The remainder of the paper is organized as follows: 
Section~\ref{sec:related_work} reviews related works. 
Section~\ref{sec:laa-x} formalizes the vulnerability concept and introduces the LAA-X framework, including LAA-Net and LAA-Former.
Section~\ref{sec:experiment} reports the experiments and discusses the results. 
Finally, Section~\ref{sec:conclu} concludes this work and suggests future investigations.

\section{Related Works on Deepfake Detection}
\label{sec:related_work}
In this section, we present an overview of previous works on deepfake detection. Specifically, we categorize them according to the type of neural architecture on which they rely, namely CNN-based and ViT-based methods.

\subsection{CNN-based Deepfake Detection}
\label{subsec:cnn_works}

Earlier methods generally formulate deepfake detection as a purely binary classification~\cite{ff++, mesonet, fake_spotter, f3-net, capsulenet} using a CNN backbone, leading to poor generalization capabilities.
To address this challenge, a wide range of strategies has been investigated~\cite{caddm, qual_agnostic, ete_recons, ucf, sbi, fxray, diffusionfake} such as disentanglement learning~\cite{ete_recons, ucf}, multi-task learning~\cite{sladd, cstency_learning, localRL, content_removal, laa_net}, pseudo-fake synthesis either in the spatial domain~\cite{sbi, fxray, cstency_learning, ost} or in the frequency domain~\cite{debias_deepfake, FreqBlender}.

Despite their great potential, the aforementioned models are less robust when considering HQ deepfakes.  Indeed, these SOTA methods mainly employ traditional DNN backbones such as ResNet~\cite{resnet}, XceptionNet~\cite{xception}, and EfficientNet~\cite{efn_net}. Hence, through their successive convolutional layers, they implicitly generate global semantic features. As a result, low-level cues that can be highly informative might be unintentionally ignored, leading to poor detection performance of HQ deepfakes. It is, therefore, crucial to design adequate strategies for modeling more localized artifacts.

Alternatively, some attention-based methods such as~\cite{multi-attentional, sfdg} have been proposed. Specifically, they have made attempts to integrate attention modules to implicitly focus on low-level artifacts~\cite{multi-attentional, sfdg}. Unfortunately, the two aforementioned methods make use of a unique binary classifier trained with both real and fake images. This means that they do not consider any generalization strategy, such as pseudo-fake generation or multi-task learning. Consequently, as demonstrated experimentally, they do not generalize well to unseen datasets in comparison to other recent techniques~\cite{sbi, aunet, cstency_learning, caddm}.

\subsection{Transformer-based Deepfake Detection}
\label{subsec:vit-work}

Plain ViTs~\cite{ViT, deit} have recently attracted significant interest from the research community, demonstrating strong performance across various computer vision tasks~\cite{ViT, detr, vitpose, mae, deit}, including the general topic of image classification. Inspired by their success, numerous transformer-based deepfake detection methods~\cite{efn_vit, conv_pooling_transformer, cnn_ViT_distill, fdftnet, cnn_vit_video, xcep_vit_increLearning, FAViT, dfdt, m2tr, ict, shallow_ViT, tall_swin, self_cnn_vit, istvt} have been introduced in the literature. A representative line of work~\cite{efn_vit, conv_pooling_transformer, cnn_ViT_distill, fdftnet, cnn_vit_video, xcep_vit_increLearning, HCiT, self_cnn_vit} designs hybrid architectures that combine CNNs and ViTs simultaneously. While the CNN extracts high-level local feature maps, the ViT models long-range correlations for authenticity classification. Despite their proven performance under in-dataset evaluation settings, they typically overfit specific artifact types present in the training set, as they rely solely on vanilla binary supervision using a fixed dataset. Consequently, their generalizability to unseen manipulations remains unsatisfactory. 
To alleviate this issue, many studies have employed strategies to encourage the network to learn more generic feature representations, such as data synthesis~\cite{ict}, adaptive learning~\cite{forensicsadapter, FAViT, GiantShoulder, UDD, plug_play} based on large pretrained foundation models~\cite{CLIP,lora}, and/or large-scale training datasets~\cite{ms_celeb_1m}. Despite demonstrating improved generalization, these transformer-based approaches still suffer from their patch-based architecture, which emphasizes global reasoning. Hence, they usually fail to model subtle artifacts typically characterizing  HQ deepfakes~\cite{delvingLocal, multi-attentional, sfdg, laa_net}.
To enforce the extraction of forgery artifacts at different scales while using a transformer architecture, a recent work termed DFDT~\cite{dfdt} has extracted multi-scale representations via multi-stream ViTs coupled with an implicit re-attention strategy. However, DFTD is still trained as a binary classifier using both real and deepfake images; hence, showing poor generalization capability to unseen forgeries.

\begin{figure}
    \centering
    \includegraphics[width=\linewidth]{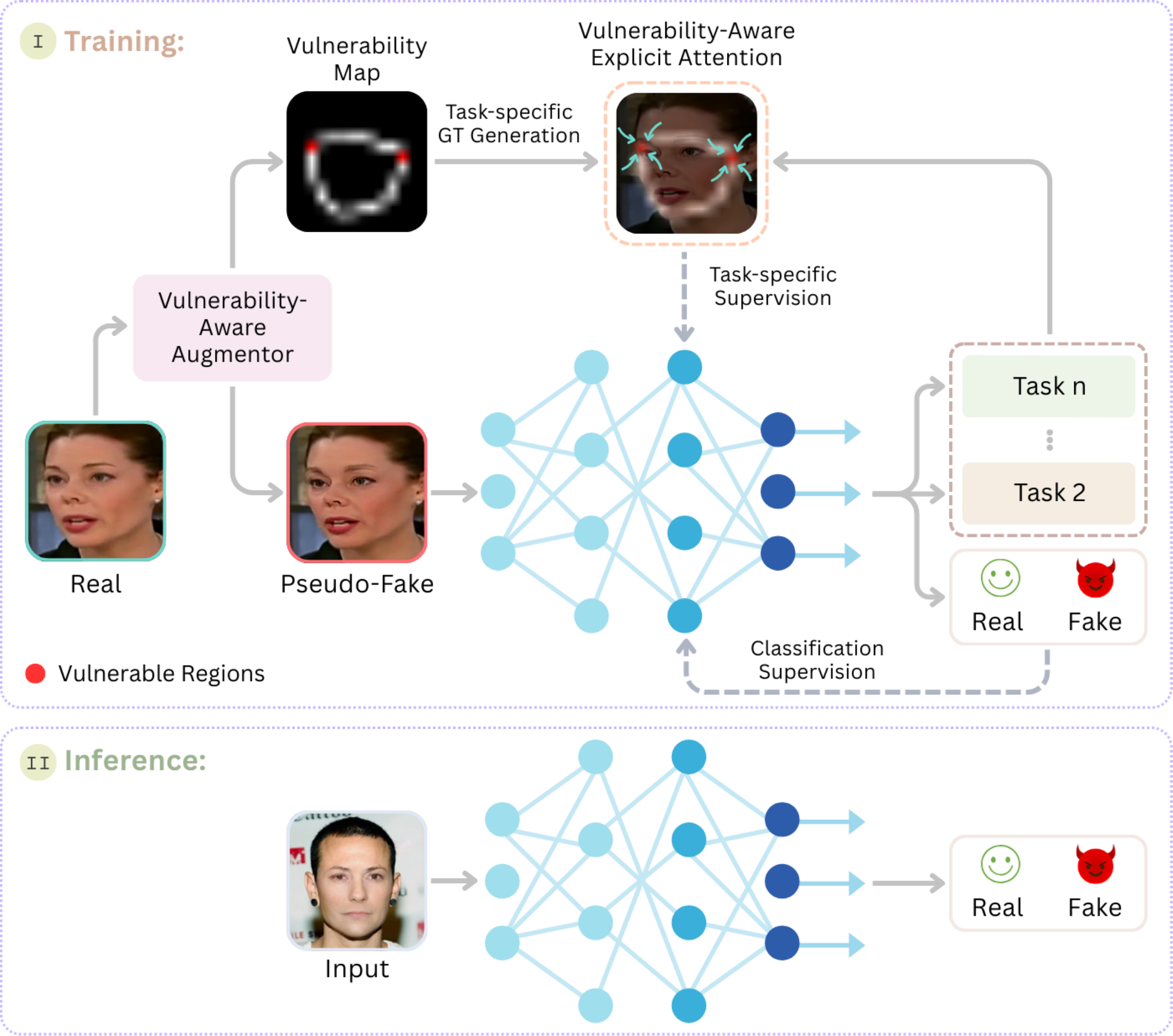}
    \caption{\textbf{Overview of the proposed LAA-X framework.} LAA-X is a multi-task learning framework that incorporates an explicit attention mechanism to vulnerable regions through the integration of generic auxiliary tasks. This strategy enables LAA-X to adequately attend to fine-grained artifact-prone areas. Particularly, these additional tasks can be removed at inference, reducing computational cost at test time.}
    \label{fig:LAA-X_overview}
    \vspace{-2mm}
\end{figure}

\section{LAA-X: A Unified Localized Artifact-Aware Attention Learning Framework}
\label{sec:laa-x}

Our goal is to introduce a method that is robust to high-quality deepfakes yet capable of handling unseen manipulations. In this section, we present a novel framework for fine-grained deepfake detection, called ``\textit{Localized Artifact Attention X (LAA-X)}''.
The main idea behind LAA-X is to enforce the model to focus on a few artifact-prone vulnerable regions in deepfakes by incorporating an explicit attention strategy through the integration of auxiliary tasks in a multi-task learning framework. By vulnerable regions, we mean the areas that are most likely to exhibit blending artifact cues. Such localized cues: (1) are common across numerous manipulation techniques; and (2) might be imperceptible to the naked eye yet present in high-quality deepfakes. As a result, detecting these vulnerable regions through an appropriate fine-grained attention mechanism simultaneously improves generalization to unseen manipulations and robustness to HQ deepfake detection. To avoid relying on specific types of deepfakes during training, compatible blending-based data synthesis strategies are proposed. Moreover, it is interesting to note that the parallel auxiliary branches are required only during training and can be removed at inference. Thus, they do not induce any additional computational cost during inference. The overview of the general LAA-X framework is shown in Figure~\ref{fig:LAA-X_overview}. 
LAA-X is architecture-agnostic as it can be adapted to both CNN and transformer backbones, yielding two different deepfake detector families. We instantiate two versions of LAA-X, including a CNN-based and a transformer-based architecture, introducing LAA-Net and LAA-Former, respectively. In Section~\ref{sec:vulnerability}, we start by defining the notion of vulnerable regions and describing their estimation within the used blending-based data synthesis techniques~\cite{sbi, fxray}. Then, Section~\ref{subsec:cnn-laa-net} and Section~\ref{subsec:transformer-laa-net} depict,  respectively, LAA-Net and LAA-Former.

\subsection{Estimation of Vulnerable Regions}
\label{sec:vulnerability}

As discussed in the previous section and illustrated in Figure~\ref{fig:LAA-X_overview}, LAA-X enforces attention to a few specific regions through additional auxiliary tasks in addition to the standard classification branch. Our hypothesis is that deepfake detection can be formulated as a fine-grained classification. Therefore, giving more attention to vulnerable regions should be an effective solution for detecting HQ deepfakes.
For that purpose, we start by defining the notion of vulnerable regions.

\begin{Definition} Vulnerable regions in a deepfake image are the areas that are more likely to carry blending artifacts. 
\end{Definition}

Depending on the architecture that is used, it is necessary to refine and extend the definition of vulnerable regions. In particular, we consider the smallest entities processed in CNN and transformer architectures, namely, pixels and patches, respectively, resulting in the following definitions.
\begin{SubDefinition} 
Vulnerable points in a deepfake image are the pixels that are more likely to carry blending artifacts. 
\end{SubDefinition}
\begin{SubDefinition}
Vulnerable patches in a deepfake image are the patches that are more likely to carry blending artifacts.
\end{SubDefinition}

\begin{figure*}
\centering
\includegraphics[width=\linewidth]{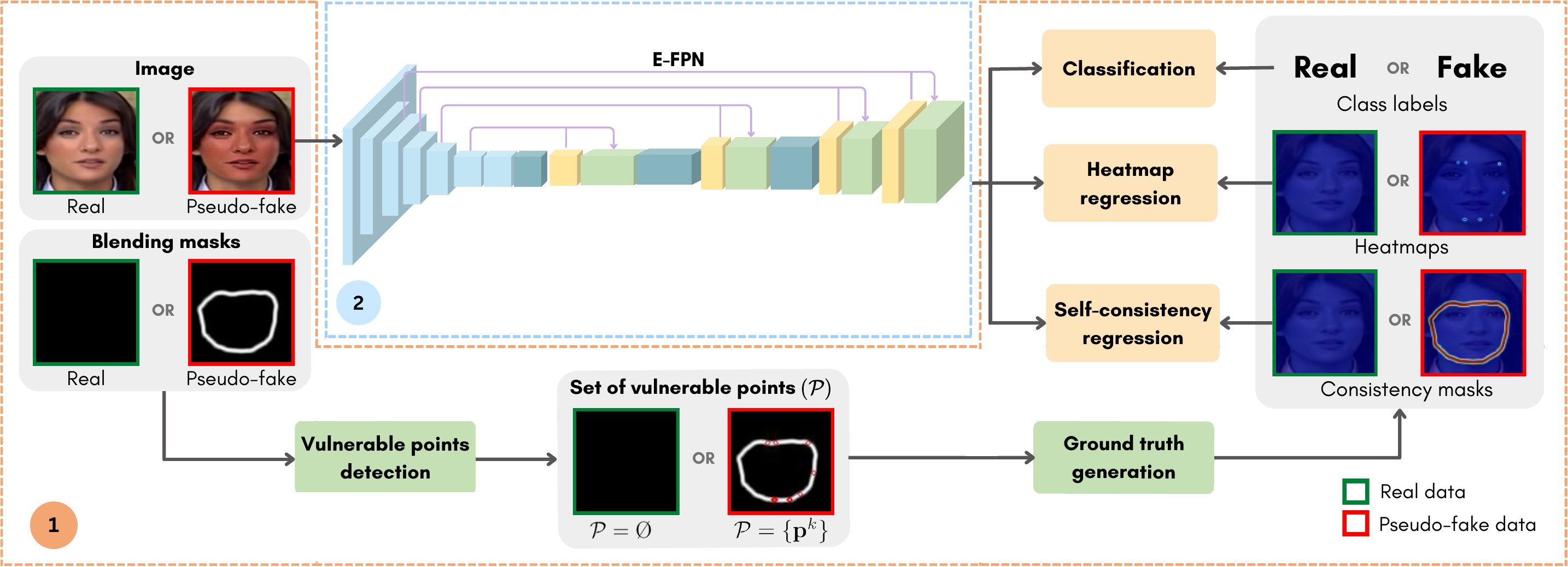}
\caption{\textbf{Overview of the proposed LAA-Net approach.} It is formed by two main components, namely, (1) an \textit{explicit attention mechanism} based on a multi-task learning framework composed of three branches, i.e., the binary classification branch, the heatmap branch, and the self-consistency branch. The heatmap and self-consistency ground-truth data are generated based on the detected vulnerable points, and (2) an \textit{Enhanced Feature Pyramid Networks (E-FPN)} that aggregates multi-scale features.}
\label{fig:wfl_ovv}
\vspace{-3mm}
\end{figure*}

\vspace{1mm}
As highlighted in~\cite{fwa, mesonet}, most deepfake generation approaches involve a blending operation for mixing the background and the foreground of two different images, $\mathbf{I}_B$ and $\mathbf{I}_F$, respectively. This implies the presence of blending artifacts regardless of the blending-based generation approach that is used.  Thus, we posit that the vulnerable regions can be seen as the areas belonging to the blending locations with the most equivalent contributions from both $\mathbf{I}_B$ and $\mathbf{I}_F$.

In this paper, we assume that we only have access to real data during training.  Hence, a blending-based data synthesis is leveraged to simulate pseudo-fakes that carry blending artifacts; hence, incorporating vulnerable regions. Such a strategy has two main advantages: (1) it avoids overfitting to specific manipulation methods seen during training, as demonstrated in several references \cite{fxray,sbi}; (2) it allows automating the estimation of ground-truth vulnerable regions to train the proposed multi-task learning framework, enabling explicit attention to those areas.

Specifically, given a foreground image and a background image denoted by $\mathbf{I_{\text F}}$ and $\mathbf{I_{\text B}}$, respectively, we adopt the blending-based synthesis method used in ~\cite{fxray,sbi} that produces a manipulated face image denoted by $\mathbf{I_{\text M}}$ as follows, 
\begin{equation}
    \mathbf I_{\text M} = \mathbf M \odot \mathbf I_{\text F} + (1-\mathbf M) \odot \mathbf I_{\text B} \ , 
    \label{equa:blending_fomular}
\end{equation}
where  $\mathbf M$ is the deformed Convex Hull mask with values varying between $0$ and $1$, and $\odot$ denotes the element-wise multiplication operator. Inspired from \cite{fxray}, a blending boundary mask $\mathbf{B}=~(b_{ij})_{i, j \in [\![ 1, D]\!]  }$ is then computed as follows, 

\begin{equation}
  \mathbf B = 4\text{ . } \mathbf{M}\odot(\mathbf{1}-\mathbf{M}) \ ,  
  \label{equa:vulner_inten_map}
\end{equation}
with $\mathbf 1$ being an all-one matrix, $D$ the height and width of $\mathbf B$, and $b_{ij}$ its value at the position $(i,j)$. 
It can be observed from Eq.~(\ref{equa:blending_fomular}) that the contributions of $\mathbf I_{\text F}$ and $\mathbf I_{\text B}$ to the pseudo-fake $\mathbf I_{\text M}$ are described by the two matrices $\mathbf{M}$ and $(\mathbf{1}-\mathbf{M})$. 
Hence, the more balanced the values of $\mathbf{M}$ and $(\mathbf{1}-\mathbf{M})$ are at a pixel $(i,j)$, the higher the value of $b_{ij}$ is, indicating a higher impact from the blending operation, and vice versa.  
Note that if an input image is real, $\mathbf B$ should be set to $\mathbf 0$. 
As such, the blending mask $\mathbf B$  is used for estimating the set of vulnerable regions denoted by $\mathcal{P}$ as follows,
\begin{equation}
\mathcal P= \argmax_{(i,j) \in [\![ 1,D_f ]\!] ^2}(f(\mathbf{B})) , 
\label{equa:set_vulnerable_regions}
\end{equation}
where  $f$ defines a sampling-aggregation strategy that is used to fit the type of architecture being considered,  $D_f$ the height and the width of $f(\mathbf{B})$, and $[\![ \textbf{ } ]\!] $ an integer interval. Note that $\mathcal{P}$ can include more than one region, since $f(\mathbf{B})$ can be maximal at several locations. We detail below how $f$ is defined for retrieving specifically vulnerable points and vulnerable patches.

\begin{figure*}[htbp]
    \centering
    \begin{minipage}{0.385\linewidth}
        \includegraphics[width=0.96\linewidth]{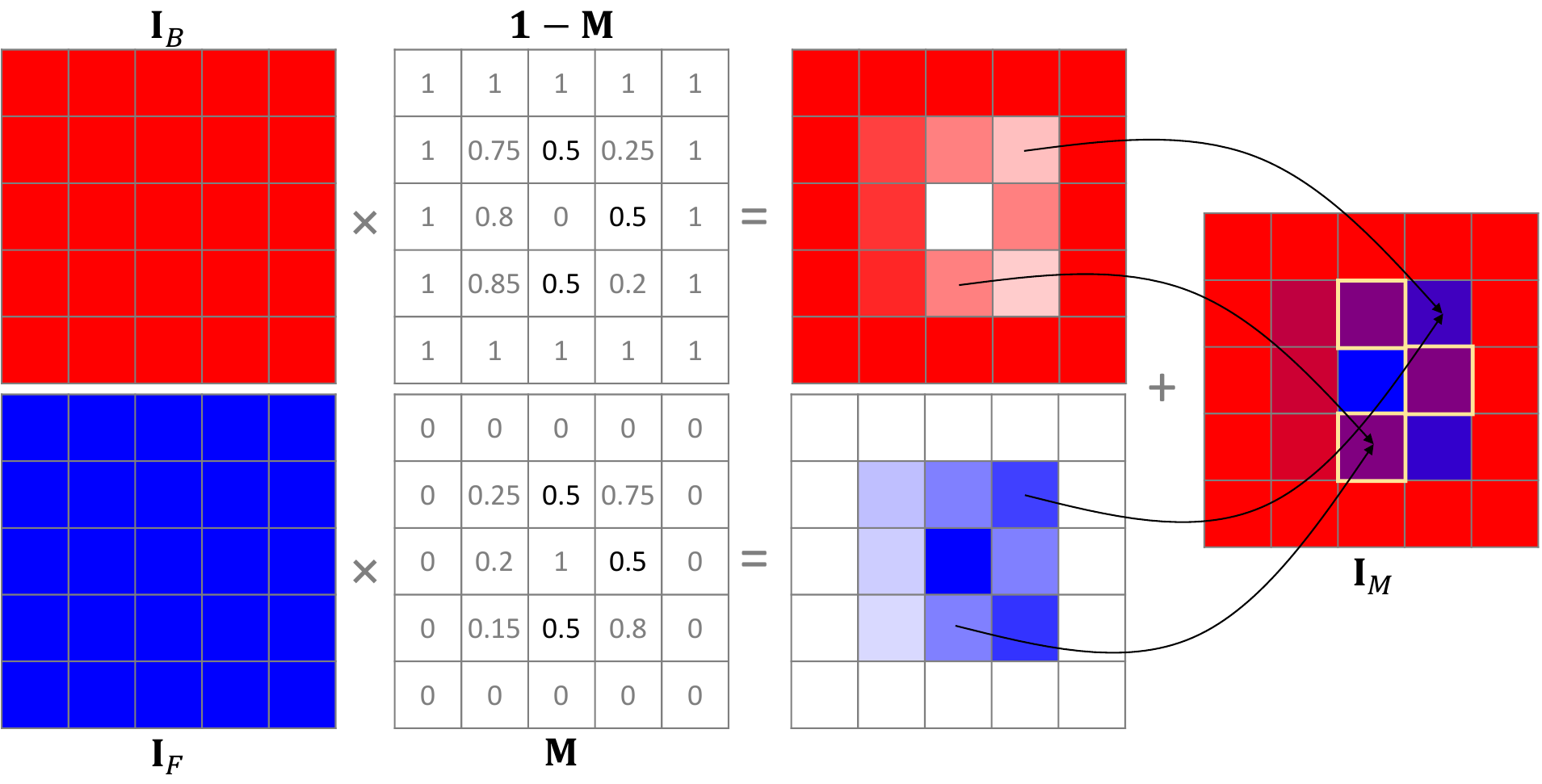}
        \caption{\textbf{Extraction of the vulnerable points.}}
        \label{fig:vulnerable_points_simple}
    \end{minipage}
    \hfill
    \begin{minipage}{0.605\linewidth}
    \centering
       \includegraphics[width=0.96\linewidth]{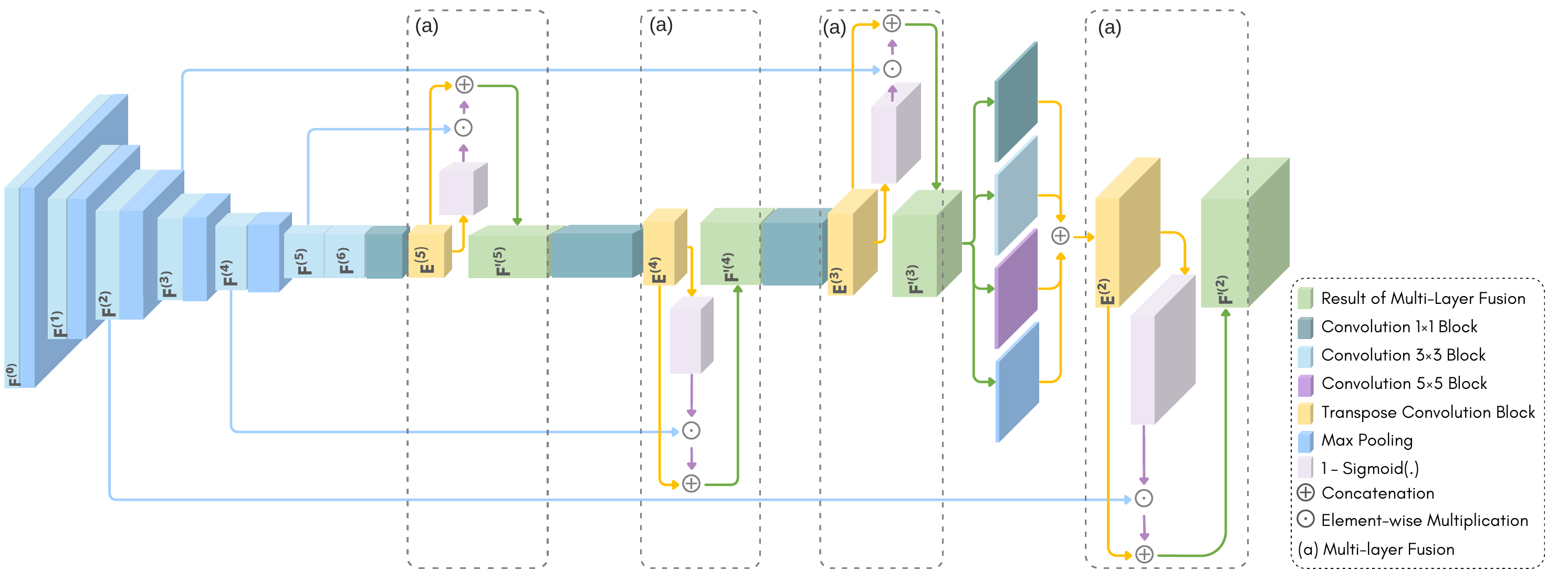}
       \caption{\textbf{Architecture of the proposed Enhanced Feature Pyramid Network (E-FPN).}}
        \label{fig:efpn}
    \end{minipage}
    \vspace{-2mm}
\end{figure*}

\vspace{1mm}
\textbf{Vulnerable Points:}\label{subsec:vulner_point} As discussed earlier, these points are compatible with CNN architectures that operate at the pixel level.  Hence, it will be used in LAA-Net in the next section. To extract vulnerable points from the blending mask $\mathbf{B}$, the function $f$ in Eq. \eqref{equa:set_vulnerable_regions} is defined as the identity function $f(\mathbf{B})=\mathbf{B}$, as no sampling or aggregation is needed. As a result, the variable $D_f$ is equal to $D$ in this context. 
Figure~\ref{fig:vulnerable_points_simple} illustrates the extraction of vulnerable points (represented as purple cells with yellow borders).

\vspace{1mm}
\textbf{Vulnerable Patches:}\label{subsec:vulner_patch}
As mentioned previously, vulnerable patches are suitable for transformer architectures that work at the patch level. Hence, $f$ is defined as the composition of two functions $f_1$ and $f_2$, as $f = f_2 \circ f_1$.
Specifically, we apply to $\mathbf B$, the patching function  $f_1$  that extracts  $N$ non-overlapping patches denoted as $\Tilde{\mathbf B}=~(\Tilde{\mathbf B}_{lm})_{l, m \in [\![ 1, \sqrt{N}]\!]}$ of dimension $P \times P$  such that $\Tilde{\mathbf B}=f_1({\mathbf B})$. 
Finally, a maxpooling operation denoted $f_2$ is employed for aggregating the information within one patch. The variable $D_f$ is therefore equal to $\sqrt{N}$ in this case. Note that other aggregation operations are considered for $f_2$ in Section~\ref{sec:experiment}.  This process is illustrated in Figure~\ref{fig:fakeformer_ovv}-III.

Even though we focus only on blending-based deepfakes, our experiments (Section~\ref{sec:experiment}) demonstrate that our models are capable of detecting various types of deepfakes, including diffusion-based ones. Extending the notion of vulnerability to non-blending artifacts is a promising direction that we will explore in future works.
In the following, we describe how the notion of vulnerable points and vulnerable patches is used within two different types of architectures, including CNNs and transformers, respectively.

\subsection{CNN-based LAA-X: LAA-Net}
\label{subsec:cnn-laa-net}

In this section, we describe the proposed CNN-based LAA-X, namely LAA-Net. An overview of LAA-Net is provided in Figure~\ref{fig:wfl_ovv}. It incorporates: (1) an \textit{explicit attention mechanism} and (2) a new architecture based on an \textit{enhanced FPN}, called \textit{E-FPN}.
First, the proposed attention mechanism aims to explicitly focus on artifact-prone pixels referred to as vulnerable points (a formal definition is given in Section~\ref{sec:vulnerability}). 
Specifically, a multi-task learning framework composed of three simultaneously parallel optimized branches, namely (a) classification, (b) heatmap regression, and (c) self-consistency regression, is introduced, as depicted in Figure~\ref{fig:wfl_ovv}. The classification branch predicts whether the input image is fake or real, while the two other branches aim to give attention to vulnerable pixels.
Second, E-FPN allows extracting multi-scale features without injecting redundancy. This enables modeling low-level features, which can better discriminate subtle inconsistencies.

\vspace{1mm}
\subsubsection{Explicit Attention to Vulnerable Points}
\label{subsec:efag}
In the following, we describe the proposed explicit attention mechanism guided by the two auxiliary branches, namely, the heatmap and the self-consistency branches, and explain how vulnerable points are utilized for training those branches.

\paragraph{Heatmap Branch}
\label{subsec:heatmapbranch}
In general, forgery artifacts not only appear in a single pixel but also affect its surroundings. Hence, considering vulnerable points as well as their neighborhoods is more appropriate for effectively discriminating deepfakes, especially in the presence of images with local irregularities caused by noise or illumination changes. To model that, we propose to use a heatmap representation that encodes at the same time the information of both vulnerable points as well as their neighbor points. 

More specifically, ground-truth heatmaps are generated by fitting an \textit{Unnormalized Gaussian Distribution} for each pixel  $\mathbf{p}^{k} = (p_x^k, p^k_y)$ $\in$  $\mathcal P$. The pixel $\mathbf{p}^{k}$  is considered as the center of the Gaussian Mask $\mathbf{G}^{k}$. To take into account the neighborhood information of $\mathbf{p}^{k}$, the standard deviation of $\mathbf{G}^{k}$ is adaptively computed. In particular, inspired by the work of~\cite{cornetnet}, the standard deviation $\sigma_{k}$ of $\mathbf{p}^{k}$ is computed based on the width and the height of the blending boundary mask $\mathbf B$ with respect to the point $\mathbf{p}^{k}$. Similar to \cite{cornetnet}, a radius $r_k$ is computed based on the size of the set of virtual objects that overlap the mask centered at $\mathbf{p}^{k}$ with an Intersection over Union (IoU) greater than a threshold $t$.
In all our experiments, we set $t$ to $0.7$ and we assume that $\sigma_k = \frac{1}{3}r_k$. Hence, $\mathbf{G}^{k}=~(g_{ij}^{k})_{i, j \in [\![ 1, D]\!]  }$ is computed as follows,

\begin{equation}
g_{ij}^{k}= e^{-\frac{(i - p_x^k)^2 + (j - p_y^k)^2}{2\sigma_{k}^2}} \ , 
\label{equa:2D_unGauss}
\end{equation}
where $i$ and $j$ refer to the pixel position.
The ground-truth heatmap $\mathbf H$ is finally constructed by superimposing the set $\mathcal G= \{ \mathbf{G}^{k}\}_{k \in  [\![ 1, \text{card}(\mathcal P)]\!]}$. A figure depicting the heatmap generation process is provided in the supplementary materials.

For optimizing the heatmap branch, the following focal loss~\cite{focal_loss} is used, 
\begin{equation}
 {L}_\text{H}= \sum_{i,j}^{D}{-(1-\Tilde{h}_{ij})^{\gamma}\log{\Tilde{h}_{ij}}} \ ,
\end{equation}
such that,

\begin{equation}
 \Tilde{h}_{ij} =
\begin{cases}
    \hat{h}_{ij} & \text{ if } h_{ij} = 1 \ , \\
    1 - \hat{h}_{ij} &  \text{ otherwise } \ , \\ 
\end{cases}
\end{equation}

\noindent with $\hat{h}_{ij}$ and $h_{ij}$ being the value of the predicted heatmap $\hat{\mathbf{H}}$ and the ground-truth $\mathbf{H}$ at the pixel location $(i,j)$, respectively. The hyperparameter $\gamma$ is used to stabilize the adaptive loss weights.

\paragraph{Self-consistency Branch} 

To enhance the proposed attention mechanism, the idea of learning self-consistency proposed in~\cite{cstency_learning} is revisited to fit our context. Instead of computing the consistency values for each pixel of the mask, we consider only the vulnerable location. Since the set $\mathcal P$ might include more than one pixel (the blending mask can include several pixels with equal maximum values), we randomly choose one of them, which we denote by $\mathbf p^s$, for generating the self-consistency ground-truth matrix. Hence, the generated matrices denoted by $\mathbf{C}$ are $2$-dimensional and not $4$-dimensional as in the original method. Given the randomly selected vulnerable point $\mathbf p^s=(u,v)$, the self-consistency $\mathbf C$ matrix is computed as, 
\begin{equation}
\mathbf C = \mathbf 1 - |b_{uv}.\mathbf 1 - \mathbf{B}| \ ,
\label{eq:consistency_gt}
\end{equation}
where $|.|$ refers to the element wise modulus and $\mathbf 1$ is an all-one matrix.

This refinement allows for reducing the model size and, consequently, the computational cost. It can also be noted that even though our method is inspired by~\cite{cstency_learning}, our self-consistency branch is inherently different. In~\cite{cstency_learning}, the consistency is calculated between the foreground and background, whereas we measure the consistency between the vulnerable point and the other pixels of the blended mask. 
The self-consistency loss $L_\text{C}$ is then computed as a binary cross entropy loss between $\mathbf C$ and the predicted self-consistency~$\hat{\mathbf C}$.

\paragraph{Training Strategy} The LAA-Net is optimized using the following loss,

\begin{equation}
    L = {L}_{\text{BCE}} + \lambda_1 {L}_{\text{H}}  + \lambda_2 {L}_{\text{C}} \ ,
\label{equa:total_loss} 
\end{equation}

\noindent where ${L}_{\text{BCE}}$ denotes the binary cross-entropy classification loss. ${L}_{\text{H}}$ and ${L}_{\text{C}}$ are weighted by the hyperparameters $\lambda_1$ and $\lambda_2$, respectively. Note that only real and pseudo-fakes are used during training.

\vspace{1mm}
\subsubsection{Enhanced Feature Pyramid Network (E-FPN)}
\label{subsec:efpn}
Feature Pyramid Networks (FPN) are widely adopted feature extractors capable of complementing global representations with multi-scale low-level features captured at different resolutions~\cite{fpn_obdet}. This makes them ideal candidates for implicitly supporting the heatmap and self-consistency branches towards fine-grained deepfake detection. Although some attempts have been made to exploit multi-scale features~\cite{caddm}, no previous works have considered FPN in the context of deepfake detection. 

Over the last years, several FPN variants have been proposed for numerous computer vision tasks~\cite{focal_loss, fcos, fpn_seg, fpn_obdet}. Nevertheless, these FPN-based methods usually lead to the generation of redundant features, which might, in turn, lead to the overfitting of the model~\cite{ayinde2019regularizing}. Moreover, as described in Section~\ref{sec:intro}, small discrepancies are gradually eliminated through the successive convolution blocks~\cite{multi-attentional}, going from high-resolution low-level to low-resolution high-level features. Consequently, the last block outputs usually contain global features where local artifact-sensitive features might be discarded. To overcome this issue, we introduce a new alternative referred to as Enhanced Feature Pyramid Network (E-FPN) that is integrated in the proposed LAA-Net architecture. The E-FPN goal is to propagate relevant information from high to low-resolution feature representations.

As shown in Figure~\ref{fig:efpn}, we denote the output shape of the $N-1$ latest layers by $(n^{(l)}, D^{(l)}, D^{(l)})$  with $l$ $\in$ $ [\![ 2,N]\!]$. For the sake of simplicity, we assume that the shape of the feature maps is square. For a given layer $l$, $n^{(l)}, D^{(l)}$ and $\mathbf F^{(l)}$ correspond, respectively, to its feature dimension, its height and width, and its output features. For strengthening the textural information in the ultimate layer $\mathbf F^{(N)}$, we propose to take advantage of the features generated by previous layers $\mathbf F^{(l)}$ with $l~\in$~$ [\![ 2, N-1]\!]$. Concretely, for each layer $l$, a convolution followed by a transpose convolution is applied to $\mathbf F^{(l+1)}$. The obtained features are denoted by $\mathbf E^{(l)}$ and have the same shape as $\mathbf F^{(l)}$. Then, a sigmoid function is applied to $\mathbf E^{(l)}$ returning probabilities. The latter indicates the pixels that contributed to the final decision. For enriching $\mathbf F^{(l+1)}$ while avoiding redundancy related to the most contributing pixels, the features $\mathbf F^{(l)}$ are filtered by computing $(1 - \text{sigmoid}(\mathbf E^{(l)}))^{\gamma_w}$ resulting in a weighted mask. The latter is concatenated along the same axis with $\mathbf E^{(l)}$ for obtaining the final features.
This operation is iterated for all the layers $l\in$ $[\![ 2, N-1]\!]$. In summary, the final representation $\mathbf  F'^{(l)}$  is obtained as follows,
\begin{equation}
   \mathbf  F'^{(l)}= (\mathbf F^{(l)} \odot  (1 - \mathrm{sigmoid}(\mathbf  E^{(l)}))^{\gamma_w}\oplus \mathbf  E^{(l)}) \ ,
\end{equation}
where $\mathbf  E^{(l)}=\mathfrak{T}(f(\mathbf F'^{(l+1)})$ with $\mathbf F'^{(l+1)} = \mathbf F^{(l+1)}$ if $l = N-1$, such that $f$ and $\mathfrak{T}$,  are respectively the convolution and transpose convolution operators, and $\oplus$ refer to the concatenation operator. The hyper-parameter $\gamma_w$ is set to $1$ in all our experiments. The relevance of E-FPN in the context of deepfake detection is experimentally demonstrated in Section~\ref{sec:experiment}, as compared to the traditional FPN.

\subsection{Transformer-based LAA-X: LAA-Former}
\label{subsec:transformer-laa-net}

\begin{figure*}
    \subfloat[]{
        \centering
        \includegraphics[width=0.38\textwidth]{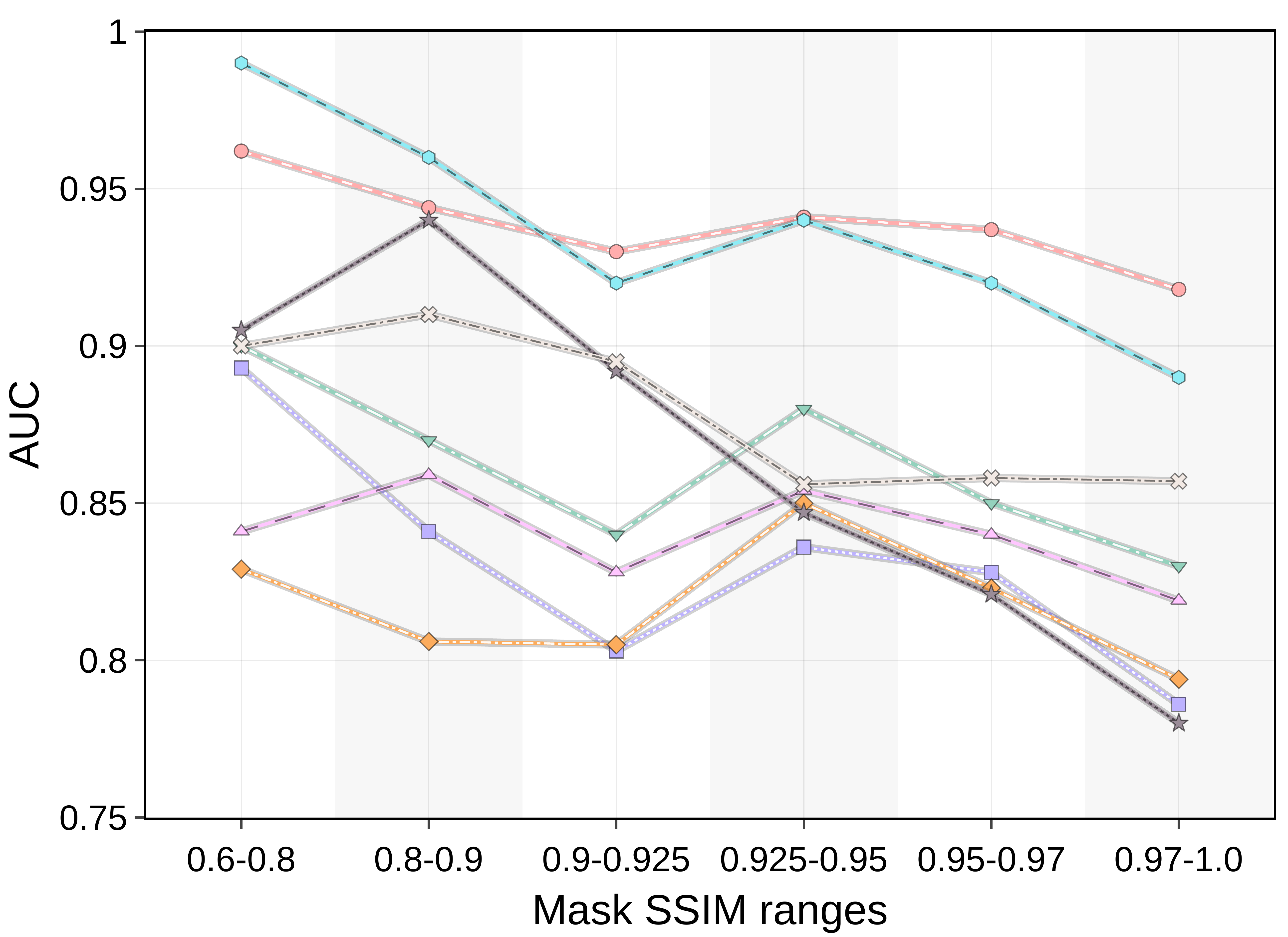}
        \label{fig:ssim_ranges}
    }
    \hfill
    \subfloat[]{
        \centering
        \includegraphics[width=0.61\textwidth]{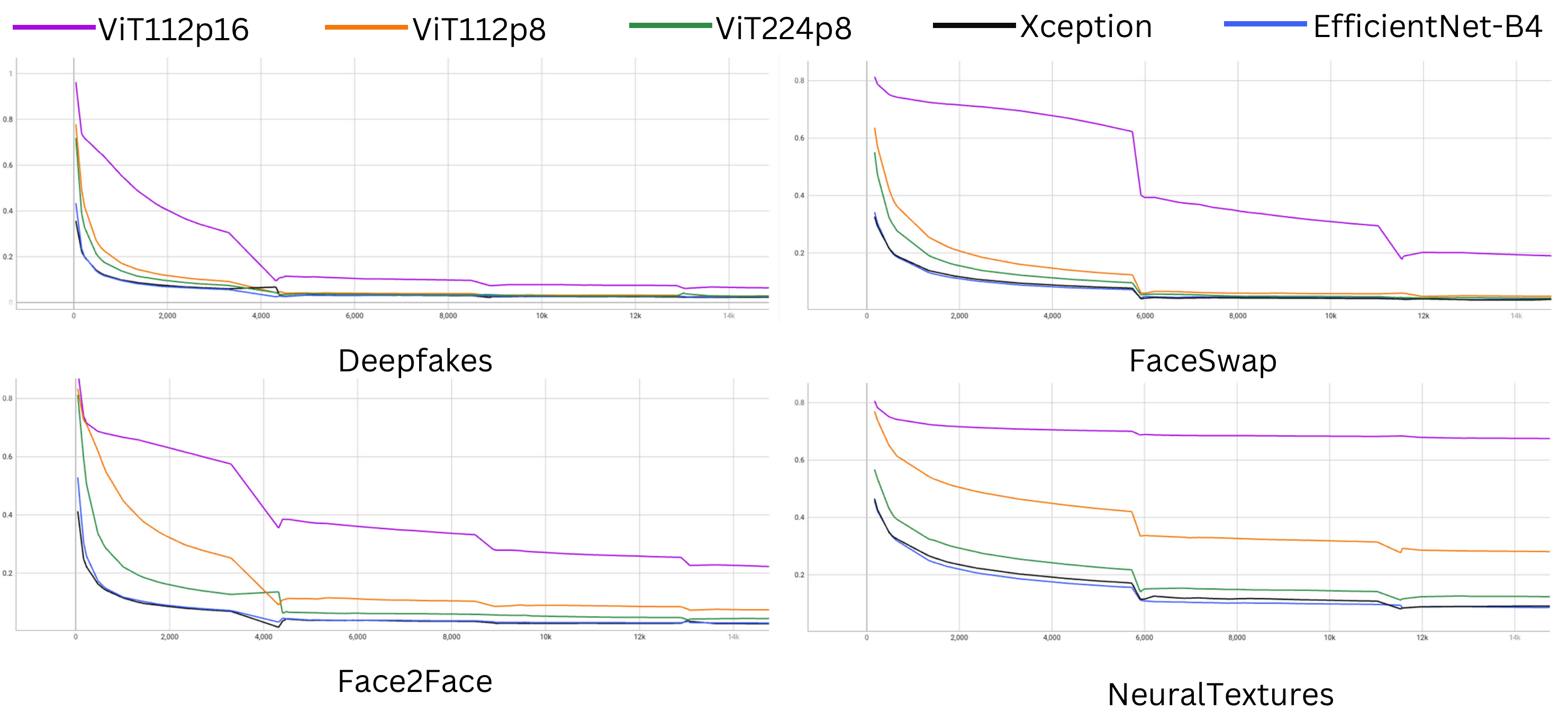}
        \label{fig:ViT_CNN_losses}
    }
    \caption{\textbf{Experiments to analyze the capability of transformer-based networks in deepfake detection.} \textit{(a)} Generalization performance comparison of baseline classifiers (ViT\cite{ViT}+SBI\cite{sbi}(\textcolor{ViT}{$\bigcdot$}), Swin\cite{swin}+SBI\cite{sbi}(\textcolor{Swin}{$\bigcdot$})) with two specialized transformer-based (FAViT~\cite{FAViT}(\textcolor{FAViT}{$\bigcdot$}), ForensicsAdapter~\cite{forensicsadapter}(\textcolor{ForensicsAdapter}{$\bigcdot$})) and four CNN-based methods (LAA-Net\cite{laa_net}+SBI\cite{sbi}(\textcolor{LAA-Net}{$\bigcdot$}), CADDM\cite{caddm}(\textcolor{CADDM}{$\bigcdot$}), EfficientNet\cite{efn_net}+SBI\cite{sbi}(\textcolor{EFN_SBI}{$\bigcdot$}), Xception~\cite{xception}+SBI~\cite{sbi}(\textcolor{Xception_SBI}{$\bigcdot$})) across different ranges of Mask-SSIM~\cite{mssim_pose}. 
    All methods adopt FF++~\cite{ff++} for the training and are tested cross-dataset on CDF2~\cite{celeb_df}. More details are described in Section~\ref{subsec:ViT_CNN_qual_DF}.
    \textit{(b)} Evolution of the training loss of ViT under different configurations (variation of input resolution and patch size), Xception and EfficientNet, across four types of deepfakes in FF++~\cite{ff++}.}
    \vspace{-4mm}
\end{figure*}

\begin{figure*}
    \centering
    \includegraphics[width=\linewidth]{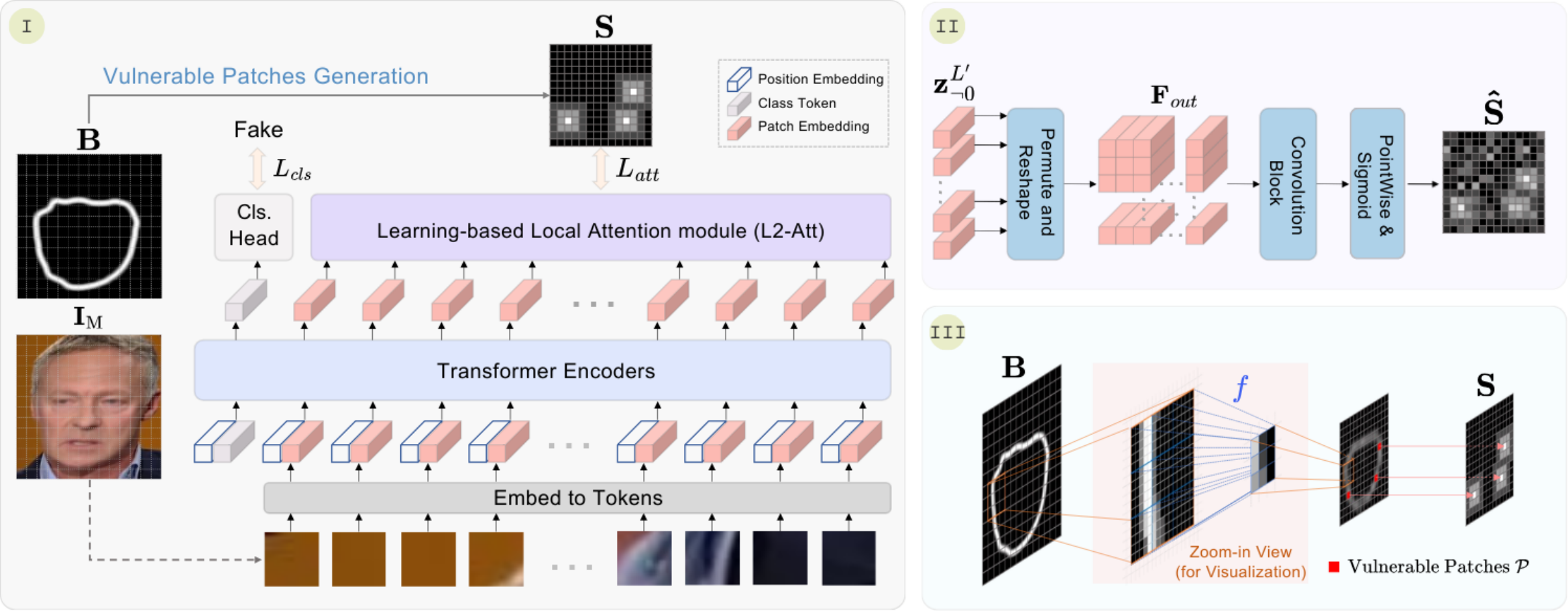}
    \caption{\textbf{The proposed Transformer-based LAA-X method.} \textit{(I)} The overall LAA-Former framework, \textit{(II)} the L2-Att module, and \textit{(III)} the ground-truth generation of vulnerable patches.}
    \label{fig:fakeformer_ovv}
    \vspace{-3mm}
\end{figure*}

As discussed in Section~\ref{sec:intro}, LAA-Net primarily captures local dependencies with limited capabilities for reasoning over spatially distant regions, which are often interrelated in facial images.
While extracting localized features is crucial~\cite{delvingLocal, multi-attentional, sfdg}, modeling the relationship between different regions can provide complementary information for more accurate detection.
Therefore, we propose to generalize the explicit attention mechanisms driven by vulnerable regions to transformers, resulting in LAA-Former. 
An overview of the proposed approach is illustrated in Figure~\ref{fig:fakeformer_ovv}-I, consisting of a plain ViT coupled with a lightweight module that enforces the model to predict the locations of vulnerable patches. Hence, this module, called ``Learning-based Local Attention (L2-Att)'', complements the well-defined implicit self-attention mechanism of the ViT.
It is noted that we refer to the transformer backbone as ViT for the sake of simplicity. However, our method is also compatible with other transformer-based architectures, such as Swin transformers~\cite{swin}, as demonstrated in Section~\ref{sec:experiment}.
Similar to LAA-Net, LAA-Former is trained using only real and pseudo-fake data.

In the following, we first investigate the specific challenges associated with the use of plain ViTs in deepfake detection (Section~\ref{sec:investigation} ). Accordingly, we present the proposed explicit attention module L2-Att that aims to improve the performance of ViT in the context of deepfake detection (Section~\ref{subsec:l2-att} ).

\vspace{1mm}
\subsubsection{ViT and Deepfake Detection: Where is the Gap?}
\label{sec:investigation}
We start by introducing our primary hypothesis: unlike CNNs, ViTs focus more on global representations~\cite{ViT, twin, nat, swin, PVT}, given their patch-based architecture. Consequently, they struggle to effectively capture local features~\cite{twin, nat} that are crucial for identifying subtle artifacts in HQ deepfakes~\cite{delvingLocal, multi-attentional, sfdg}.
The plausibility of this assumption is investigated by conducting the following experiments described below.

\paragraph{Generalization performance with respect to the quality of deepfakes} 
\label{subsec:ViT_CNN_qual_DF}
Here, our goal is to quantify the detection capabilities of ViTs as compared to CNNs with respect to the quality of the encountered deepfakes. To that aim, we compare in Figure~\ref{fig:ssim_ranges} the performance of conventional transformers (plain ViT~\cite{ViT} and Swin~\cite{swin}),
transformer-based methods specifically tailored for deepfake detection (FAViT~\cite{FAViT}, ForensiscAdapter~\cite{forensicsadapter})
and
vanilla CNN architectures (EfficientNet~\cite{efn_net}, Xception~\cite{xception}) and CNN-based methods specialized in deepfake detection (LAA-Net~\cite{laa_net}, CADDM~\cite{caddm}) across different ranges of Mask-SSIM~\cite{mssim_pose} on the CDF2~\cite{celeb_df} dataset. 
All methods rely on FF++~\cite{ff++} for training and testing on CDF2~\cite{celeb_df}, following the standard cross-dataset protocol~\cite{fxray, sladd, sfdg, cstency_learning}.
For a fair comparison, we clarify that we train ViT, Swin, EfficientNet, and Xception with the same data synthesis method, i.e., SBI~\cite{sbi} which is also used in SBI~\cite{sbi} \footnote{LAA-Net, CADDM, and SBI use EfficientNet-B4 as the default backbone.} while CADDM and ForensicsAdapter are trained with their data synthesis algorithms.
The performance of CADDM, ForensicsAdapter, and FAViT is reproduced using the official pretrained weights.

It can be observed from Figure~\ref{fig:ssim_ranges} that the performance of ViT is relatively good for low Mask-SSIM values, but drops more importantly at higher values as compared to other methods. The performance of Swin, on the other hand, does not deteriorate as much, potentially due to its ability to extract low-level local features through its local window design; however, it still shows less stability than LAA-Net and ForensicsAdapter. Notably, FAViT, which combines ViT with a CNN via an \textit{implicit} local-attention scheme, exhibits better capabilities than standard CNNs, ViT, and even the specialized CNN-based CADDM at higher SSIM ranges. However, since the method relies on specific deepfakes during training, it tends to achieve poor generalization when encountering unseen generation methods (i.e., from FF++ $\rightarrow$ CDF2). As such, these observations support our hypothesis.

\begin{figure}
    \centering
    \includegraphics[width=\linewidth]{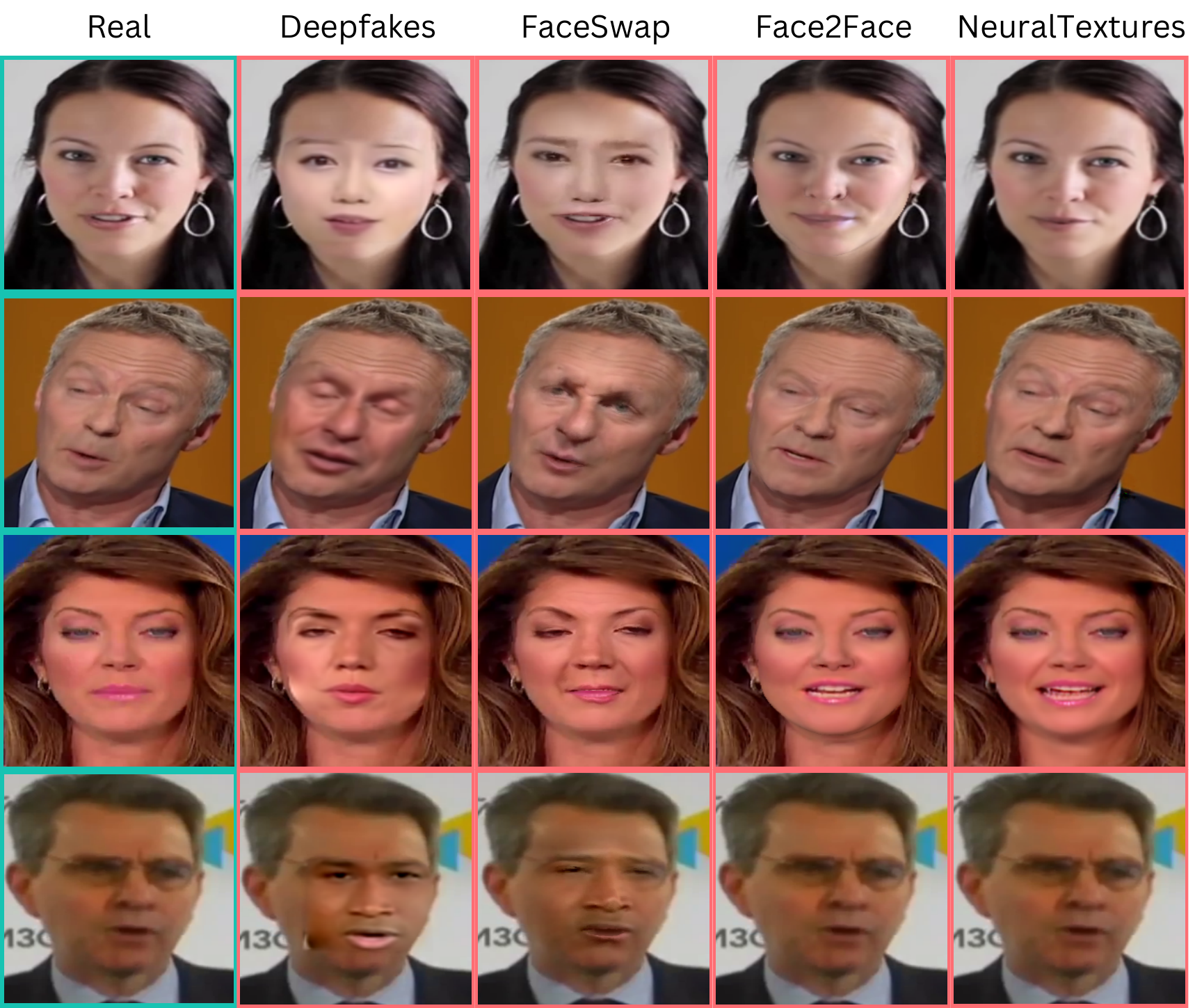}
    \caption{\textbf{Examples are randomly selected to illustrate four types of deepfakes in the common FF++~\cite{ff++} dataset.} It includes Deepfakes (DF)~\cite{deepfake}, FaceSwap (FS)~\cite{faceswap}, Face2Face (F2F)~\cite{face2face}, and NeuralTextures (NT)~\cite{neutex}. }
    \label{fig:locality_levels}
    \vspace{-2mm}
\end{figure}

\paragraph{Performance of ViTs with respect to the patch size and the type of deepfakes}
\label{subsec:patchsize_DFs}
We further investigate whether there exists a \textit{correlation} between the patch size and the performance of ViT in deepfake detection. Specifically, we anticipate that smaller patch sizes would help capture more localized artifacts. For that purpose, we train a plain ViT with several configurations by varying both the patch size and the input resolution. Figure~\ref{fig:ViT_CNN_losses} depicts the evolution of the training loss through epochs. The notation ViT$X$p$Y$ in Figure~\ref{fig:ViT_CNN_losses} denotes an input resolution of $X$ with a patch size of $Y$. 
We also compare ViT to two widely-adopted CNNs, i.e.,
Xception and EfficientNet in this study. 
Both ViT and CNNs are trained on four types of deepfakes in FF++~\cite{ff++}: Deepfakes (DF)~\cite{deepfake}, FaceSwap (FS)~\cite{faceswap}, Face2Face (F2F)~\cite{face2face}, and NeuralTextures (NT)~\cite{neutex} as shown in Figure~\ref{fig:locality_levels}. For the training setups, following the conventional splits~\cite{ff++}, we train all models for $50$ epochs, using $128$ and $32$ frames uniformly extracted from each video for training and validation, respectively. 
Hence, there are a total of $460800$ and $22400$ frames for each corresponding task.
More details, e.g., optimizer, learning-rate scheduler, etc., are provided in the supplementary materials.

As F2F and NT correspond to face reenactment manipulations while FS and DF represent face-swap approaches, it is more likely to observe more subtle artifacts in F2F and NT. When comparing ViT$112$p$16$ and ViT$112$p$8$, it can be seen that a ViT with smaller patches exhibits faster convergence compared to those with larger ones. Moreover, increasing the input resolution while conserving the same patch size (see ViT$112$p$8$ and ViT$224$p$8$) contributes to the amplification of the local information encoded in each patch, also resulting in faster convergence. In both cases, this convergence gap is even more pronounced for more subtle deepfake types such as F2F and NT, indicating the importance of locality in detecting deepfakes with subtle inconsistencies. However, reducing the patch size leads to a \textit{quadratically} increasing complexity (i.e., the FLOPs are $2.2$G, $8.4$G, and $33.6$G for ViT$112$p$16$, ViT$112$p$8$, and ViT$224$p$8$, respectively). Moreover, it is worth noting that CNNs converge more rapidly compared to ViTs under different setups (i.e., even with the smallest patch size). 
This also highlights the fact that CNNs can extract local features more effectively, supporting further our hypothesis. 

Hence, we posit that by proposing a mechanism that allows focusing on subtle artifact-prone regions, we can enhance the performance of ViT for the task of deepfake detection. While some attempts have been made to introduce local ViTs such as Swin~\cite{swin}, we argue that this remains insufficient for effectively detecting deepfakes. As demonstrated for CNNs~\cite{laa_net}, implicitly incorporating local features does not guarantee that artifact-prone regions are effectively considered, highlighting the need to introduce attention strategies for explicitly focusing on localized artifacts.

\vspace{1mm}
\subsubsection{Explicit Attention to Vulnerable Patches}
\label{subsec:l2-att}

In light of the observations made in Section~\ref{sec:investigation}, we propose to inject a lightweight local attention head that we call L2-Att within ViT, resulting in LAA-Former.  The latter aims to enforce the model to focus vulnerable patches. In what follows, we depict the different components of FakeFormer.

\paragraph{Vision Transformer (ViT)}
\label{subsec:overview}

Given an image $\mathbf X \in \mathbb{R}^{C \times H \times W}$ as input, we first reshape it into a sequence of non-overlapping flattened 2D patches, denoted as $\{\mathbf x_i \in \mathbb R^{C.P^2} \text{ with } i \in [\![ 1, N]\!] \}$, where $(H, W)$ represents the input resolution, $C$ denotes the number of channels, $P \times P$ indicates the size of an image patch, and $N = \frac{H \times W}{P^2}$ denotes the number of patches. The ViT linearly maps each $\mathbf x_i$ into a patch embedding $\mathbf z^0_{i} \in \mathbb R^D$ using a learnable matrix $\mathbf E \in \mathbb R^{(C.P^2) \times D}$. Subsequently, a learnable embedding $\mathbf x_{cls} \in \mathbb R^D$ is prepended at the zero-index of embeddings $\mathbf z^0$ for the classification. Additionally, we use a learnable positional embedding $\mathbf E^{pos}$  to incorporate the position information of patches. The aforementioned process is described as follows,
\begin{equation}
    \mathbf z^0 = [\mathbf x_{cls}; \mathbf x_1\mathbf E; \mathbf x_2\mathbf E; \cdots; \mathbf x_N\mathbf E] + \mathbf E^{pos},
\label{equa:patch_embed}
\end{equation}
where $\mathbf E^{pos} \in \mathbb R^{(N+1) \times D}$ and $\mathbf z^0 \in \mathbb R^{(N+1) \times D}$. Afterward, $\mathbf z^0$ is fed into several transformer encoder blocks. Similar to ViT~\cite{ViT}, LAA-Former has $L$ blocks, each one containing a multi-head self-attention (MHSA), Layernorm (LN), and a multi-layer perceptron (MLP). The feature extraction process is described as follows,
\begin{flalign}
    \mathbf z^l &= \mathrm{MHSA}(\mathrm{LN}(\mathbf z^{l - 1})) + \mathbf z^{l - 1}, \\\nonumber
    \mathbf z^{l'} &= \mathrm{MLP}(\mathrm{LN}(\mathbf z^l)) + \mathbf z^l, 
\end{flalign}
with $l \in [\![ 1, L]\!]$ and $\mathbf z^l, \mathbf z^{l'} \in \mathbb R^{(N+1) \times D}$.
The extracted feature from the classification embedding $\mathbf z^{L'}_{0}$ is processed by a classification head composed of an MLP, resulting in the predicted category output $\hat{\mathbf y}$. In the task of deepfake detection, the categories consist of \textit{real} or \textit{fake}.

\paragraph{Learning-based Local Attention Module (L2-Att)}
\label{subsec:lgam}
Our hypothesis is that since the patch size can be too large relative to the area of artifacts, the features encoded within a patch embedding may hold insufficient information about them. Consequently, the implicit SA mechanism might overlook or miss important patches, as those containing forgeries can appear too analogous to those without. 
As also highlighted in previous work~\cite{forensicsadapter}, the blending boundary forgery only occupy a small portion of the image, naively training using standard classification loss can easily be influenced by the non-boundary areas, leading to suboptimal results.
Therefore, we propose an explicit attention mechanism to ensure that the model pays more attention to these critical patches. 
To this end, by enforcing the model to predict the locations of vulnerable patches, L2-Att can play a complementary role to the SA, strengthening the detection capability of the whole framework.

\vspace{1mm}
\textbf{Ground-Truth Generation for L2-Att.} To obtain the ground-truth to be compared to the output of L2-Att denoted as $\mathbf S$, we generate a weighted map $\mathbf S^q$ for each element $\mathbf p^q = (p_x^q, p_y^q) \in \mathcal P$ (Eq.~(\ref{equa:set_vulnerable_regions})). To take into account the neighborhood patches beneficial to consolidate the network detection, we use an unnormalized Gaussian distribution to calculate $\mathbf S^q$ as follows,
\begin{equation}
    \mathbf S^q(l,m) = e^{-\frac{(l - p^q_x)^2 + (m - p^q_y)^2}{2\sigma^2}},
    \label{equa:gauss_weighted_map}
\end{equation}
where $(l,m) \in [[1, \sqrt{N}]]$ represents the spatial position, and the standard deviation $\sigma$ is fixed to $1$ by default. We obtain $\mathbf S$ by overlaying $ \{\mathbf S^q \}_{q \in [[1, \text{card}(\mathcal P)}]]$. The ground truth generation process is also illustrated in Figure~\ref{fig:fakeformer_ovv}-III. It can be noted that, for real data, $\mathbf S$ is set to a zero matrix. 

\vspace{1mm}
\textbf{Architecture Design.}
To predict the locations of vulnerable patches, L2-Att first takes the patch embeddings $\mathbf z^{L'}_{\neg 0} \in \mathbb R^{N \times D}$ as input and processes them to produce spatial features as follows,
\begin{flalign}
    \mathbf F &= \mathrm{Permute}(\mathbf z^{L'}_{\neg 0}), \hspace{3mm} \mathbf F \in \mathbb R^{D \times N} \text{,} \\\nonumber
    \mathbf F_{out} &= \mathrm{Reshape}(\mathbf F), \hspace{3mm} \mathbf F_{out} \in \mathbb R^{D \times \sqrt{N} \times \sqrt{N}} \text{,}
    \label{equa:reshape_output}
\end{flalign}

After that, $\mathbf F_{out}$ is fed into a convolution block (ConvBlock) with a kernel size of ($3 \times 3$), followed by a pointwise convolution~\cite{pointwise} and a sigmoid activation. The predicted weighted heatmap denoted as $\hat{\mathbf S}$ describing the presence probability of vulnerable patches is obtained as follows,
\begin{equation}
    \hat{\mathbf S} = \sigma(\mathrm{PointWise}(\mathrm{ConvBlock}_{3\times3}(\mathbf F_{out}))),
    \label{equa:hm_output}
\end{equation}
where $\hat{\mathbf S} \in \mathbb R^{1 \times \sqrt{N} \times \sqrt{N}}$.
A detailed illustration of the L2-Att  module can be seen in Figure~\ref{fig:fakeformer_ovv}-II.

\paragraph{Training Objective}
\label{subsec:training_obj}

To train the network, we optimize two losses, namely the Binary Cross Entropy (BCE) loss for classification denoted as $L_{cls}(\hat{\mathbf y}, \mathbf y)$, and the regression loss related to the prediction of vulnerable patches locations, denoted as $L_{att}(\hat{\mathbf S}, \mathbf S)$. Therefore, the total loss $L$ is defined as follows,
\begin{equation}
    L = L_{cls} + \lambda_{att} L_{att} \text{,}
    \label{equa:total_loss_transformer}
\end{equation}
where $\lambda_{att}$ is a balancing factor between the two losses. Similarly in LAA-Net, we employ the focal loss~\cite{focal_loss} to compute $L_{att}(\hat{\mathbf S}, \mathbf S)$. 

\section{Experiments}
\label{sec:experiment}
In this section, we start by presenting the experimental settings. Then, we compare the performance of LAA-X to SOTA methods, both qualitatively and quantitatively. Finally, we conduct an ablation study to validate the different components of LAA-X.

\subsection{Experimental Settings}
\noindent\textbf{Datasets.}\label{subsec:datasets}
The FF++~\cite{ff++} dataset is used for training and validation. In our experiments, we follow the standard splitting protocol of~\cite{ff++}. This dataset contains $1000$ original videos and $4000$ fake videos generated by four different manipulation methods, namely, Deepfakes (DF)~\cite{deepfake}, Face2Face (F2F)~\cite{face2face}, FaceSwap (FS)~\cite{faceswap}, and NeuralTextures (NT)~\cite{neutex}. In the training process, we utilize only real images to dynamically generate pseudo-fakes, as discussed in Section~\ref{sec:laa-x}. To evaluate the generalization capability of the proposed approach as well as its robustness to high-quality deepfakes, we follow the cross-dataset setting on seven challenging datasets incorporating different quality of deepfakes, namely, \textbf{Celeb-DFv2}~\cite{celeb_df} (CDF2), \textbf{Google DeepFake Detection}~\cite{dfd} (DFD), \textbf{DeepFake Detection Challenge}~\cite{dfdc} (DFDC) and its preview version (i.e., \textbf{DeepFake Detection Challenge Preview}~\cite{dfdcp} (DFDCP)), \textbf{WildDeepfake}~\cite{wdf} (DFW), a diffusion-based test set \textbf{DiffSwap}~\cite{diffusionface}, and \textbf{DF40}~\cite{df40}.
To assess the quality of the considered datasets, we compute the Mask-SSIM\footref{mssim_defi} for each benchmark. In particular, CDF2~\cite{celeb_df} is formed by the most realistic deepfakes with an average Mask-SSIM~\cite{mssim_pose, celeb_df} value of $0.92$, followed by DFD, DF40, DFDC, and DFDCP with an average Mask-SSIM of $0.88$, $0.87$, $0.84$ and $0.84$, respectively. We note that computing the Mask-SSIM for DFW and DiffSwap was not possible since real and fake images are not paired. 
Further details on these considered datasets are provided in supplementary materials.

\begin{table*}
\centering
\caption{\textbf{In-dataset and Cross-dataset evaluation} in terms of AUC (\%) and AP (\%) at the \textit{video-level} on multiple deepfake datasets. Results for comparison are directly extracted from the original papers. $\ast$ indicates our reproduced results using official pre-trained weights. \textbf{Bold} and \underline{Underlined} highlight the best and the second-best performance, respectively.}
\resizebox{\linewidth}{!}{
{\rowcolors{30}{Plum!10}{SkyBlue!20}
\begin{tabular}{c c cc c HHHH ccHH ccHH ccHH cc cc}
\toprule
\multirow{3}{*}{Method} & \multirow{3}{*}{Venue} & \multicolumn{2}{c}{Training set} & \multicolumn{21}{c}{Test set} \\
\cmidrule(lr){3-4}
\cmidrule(lr){5-25}
& & \multirow{2}{*}{Real} & \multirow{2}{*}{Fake} & FF++ & \multicolumn{4}{H}{CDF1} & \multicolumn{4}{c}{CDF2} & \multicolumn{4}{c}{DFW} & \multicolumn{4}{c}{DFD} &\multicolumn{2}{c}{DFDCP} & \multicolumn{2}{c}{DFDC} \\
\cmidrule(lr){5-6}\cmidrule(lr){10-13}\cmidrule(lr){14-17}\cmidrule(lr){18-21}\cmidrule(lr){22-23}\cmidrule(lr){24-25}
 & & & & AUC & AUC & AP & AR  & mF1  & AUC & AP & AR & mF1 & AUC & AP & AR & mF1 & AUC & AP & AR & mF1 & AUC & AP & AUC & AP \\
\midrule
\midrule
Xception$^\ast$~\cite{ff++} & ICCV'19 & $\checkmark$ & $\checkmark$ & 93.60 & 58.81 & 65.59 & 55.58 & {60.17} & 61.18 & 66.93 & 52.40 & {58.78} & 65.29 & 55.37 & 57.99 & {56.65} & 89.75 & 85.48 & 79.34 & 82.29 & 69.90 & 91.98 & 58.98 & 55.32 \\

FaceXRay (w/ BI)~\cite{fxray} & CVPR'20 & $\checkmark$ & $\checkmark$ & 99.20 & 80.58 & 73.33 & - & - & 79.50 & - & - & {-} & - & - & - & {-} & 95.40 & 93.34 & - & - & 65.50 & - & - & - \\

Multi-attentional$^\ast$~\cite{multi-attentional} & CVPR'21 & $\checkmark$ & $\checkmark$ & 95.32 & 69.14 & 74.03 & 52.70 & {61.57} & 68.26 & 75.25 & 52.40 & {61.78} & 73.56 & 73.79 & 63.38 & {68.19} & 92.95 & 96.51 & 60.76 & 74.57 & 83.81 & 96.52 & 70.05 & 67.11 \\

PCL+I2G~\cite{cstency_learning} & ICCV'21 & $\checkmark$ & $\times$ & 99.11 & \textbf{98.30} & - & - & {-} & 90.03 & - & - & {-} & - & - & - & {-} & 99.07 & - & - & - & 74.27 & - & 67.52 & - \\

RECCE$^\ast$~\cite{ete_recons} & CVPR'22 & $\checkmark$ & $\checkmark$ & 99.56 & 49.96 & 63.04 & 50.87 & {56.31} & 70.93 & 70.35 & 59.48 & 64.46 & 68.16 & 54.41 & 56.59 & {55.48} & 98.26 & 79.42 & 69.57 & 74.17 & 80.98 & 92.75 & 71.19 & 68.97 \\

SBI$^\ast$~\cite{sbi} & CVPR'22 & $\checkmark$ & $\times$ & 98.23 & 82.65 & 77.09 & 70.68 & 73.75 & 85.55 & 77.81 & 68.13 & 72.65 & 67.47 & 55.87 & 55.82 & 55.85 & 96.04 & 92.79 & 89.49 & 91.11 & 82.22 & 93.24 & 69.77 & 72.25 \\

DFDT~\cite{dfdt} & Appl.Sci.'22 & $\checkmark$ & $\checkmark$ & 97.9 & - & - & - & - & 88.3 & - & - & - & - & - & - & - & - & - & - & - & 76.1 & - & - & - \\

SFDG~\cite{sfdg} & CVPR'23 & $\checkmark$ & $\checkmark$ & 99.53 & - & - & - & {-} & 75.83 & - & - & {-} & 69.27 & - & - & {-} & 88.00 & - & - & - & 73.63 & - & - & - \\

CADDM$^\ast$~\cite{caddm} & CVPR'23 & $\checkmark$ & $\checkmark$ & 99.26 & 89.36 & 93.25 & 81.41 & 86.93 & 80.70 & 87.72 & 72.56 & 79.42 & 76.31 & 79.19 & 69.35 & 73.95 & 99.03 & \underline{99.59} & 82.17 & 90.04 & 71.00 & 95.60 & 70.33 & 70.01 \\

AUNet~\cite{aunet} & CVPR'23 & $\checkmark$ & $\times$ & 99.46 & - & - & - & {-} & 92.77 & - & - & {-} & - & - & - & {-} & \underline{99.22} & - & - & - & 86.16 & - & 73.82 & - \\

LSDA~\cite{LSDA} & CVPR'24 & $\checkmark$ & $\checkmark$ & 95.8 & - & - & - & - & 89.8 & - & - & - & 75.6 & - & - & - & 95.6 & - & - & - & 81.2 & - & 73.5 & - \\

FA-ViT~\cite{FAViT} & TCSVT'24 & $\checkmark$ & $\checkmark$ & 99.6 & - & - & - & - & 93.83 & - & - & - & \bf{84.32} & - & - & - & 94.88 & - & - & - & 85.41 & - & \underline{78.32} & - \\

UDD~\cite{UDD} & AAAI'25 & $\checkmark$ & $\checkmark$ & - & - & - & - & - & 93.1 & - & - & - & - & - & - & - & 95.5 & - & - & - & 88.1 & - & - & - \\

FreqDebias~\cite{debias_deepfake} & CVPR'25 & $\checkmark$ & $\checkmark$ & - & - & - & - & - & 89.6 & - & - & - & - & - & - & - & - & - & - & - & - & - & 77.8 & - \\

\midrule

AltFreezing~\cite{altfreezing} & CVPR'23 & $\checkmark$ & $\checkmark$ & 98.60 & - & - & - & - & 89.50 & - & - & {-} & - & - & - & {-} & 98.50 & - & - & - & 70.84 & - & 71.74 & - \\

ISTVT~\cite{istvt} & TIFS'23 & $\checkmark$ & $\checkmark$ & 99.0 & - & - & - & - & 84.1 & - & - & - & - & - & - & - & - & - & - & - & 74.2 & - & - & - \\

TALL-Swin~\cite{tall_swin} & ICCV'23 & $\checkmark$ & $\checkmark$ & 99.87 & - & - & - & - & 90.79 & - & - & - & - & - & - & - & - & - & - & - & 76.78 & - & - & - \\

FakeSTormer~\cite{fakestormer} & ICCV'25 & $\checkmark$ & $\times$ & 98.4 & - & - & - & - & 92.4 & - & - & - & 74.2 & - & - & - & 98.5 & - & - & - & 90.0 & - & 74.6 & - \\

\midrule
\midrule
LAA-Net (Ours w/ BI) & CVPR'24 & $\checkmark$ & $\times$ & \underline{99.95} & 92.46 & 95.54 & 50.0 & 65.64 & 86.28 & 91.93 & 50.0 & 64.77 & 57.13 & 56.89 & 50.12 & {{53.29}} & \textbf{99.51} & \textbf{99.80} & \textbf{95.47} & \textbf{97.59} & 69.69 & 93.67 & 71.36 & 73.02 \\ 

LAA-Former (Ours w/ BI) & - & $\checkmark$ & $\times$ & 99.23 & & & & & 90.34 & 94.90 & & & 72.62 & 75.98 & & & 93.42 & 97.49 & & & 78.71 & 96.23 & 76.84 & \textbf{80.82} \\

\midrule 
LAA-Net (Ours w/ SBI) & CVPR'24 & $\checkmark$ & $\times$ & \textbf{99.96} & \underline{93.11} & \textbf{95.64} & \textbf{89.78} & {\textbf{92.62}} & \textbf{95.40} & \textbf{97.64} & \textbf{87.71} & {\textbf{92.41}} & 80.03 & \underline{81.08} & \textbf{65.66} & {\textbf{72.56}} & 98.43 & 99.40 & 88.55 & \underline{93.64} & \underline{86.94} & \underline{97.70} & 72.43 & 74.46 \\

LAA-Former (Ours w/ SBI) & - & $\checkmark$ & $\times$ & 97.67 & & & & & \underline{94.45} & \underline{97.15} & & & \underline{81.74} & \textbf{83.72} & & & 96.12 & 98.31 & & & \textbf{96.30} & \textbf{99.50} & \textbf{78.91} & \underline{80.01} \\

\bottomrule
\end{tabular}%
}}
\vspace{-3mm}
\label{tabl:cross_auc_full_metrics_2}
\end{table*}

\begin{table}
\centering
\caption{\textbf{Comparison in terms of AUC (\%) at the \textit{frame-level}} with cross-dataset evaluation on CDF2~\cite{celeb_df}, DFDCP~\cite{dfdcp}, and DiffSwap~\cite{diffusionface}.}
\resizebox{\linewidth}{!}{
{\rowcolors{10}{Plum!10}{SkyBlue!20}
\begin{tabular}{c c cc H ccc}
\toprule
\multirow{2}{*}{Method} & \multirow{2}{*}{Venue} & \multicolumn{2}{c}{Training set} & & \multicolumn{3}{c}{Cross-dataset} \\
\cmidrule{3-4}
\cmidrule{6-8}
& & Real & Fake & & CDF & DFDCP & DiffSwap \\
\midrule
\midrule
SBI~\cite{sbi} & CVPR'22 & \checkmark & $\times$ & & 78.59 & 78.05 & 75.20 \\

CADDM~\cite{caddm} & CVPR'23 & \checkmark & \checkmark & & 73.16 & 65.19 & 75.58 \\

LSDA~\cite{LSDA} & CVPR'24 & \checkmark & \checkmark & & 83.0 & 81.5 & - \\

DiffusionFake (EFN-B4)~\cite{diffusionfake} & NeurIPS'24 & \checkmark & \checkmark & & 83.17 & 77.35 & 82.02 \\

DiffusionFake (ViT-B)~\cite{diffusionfake} & NeurIPS'24 & \checkmark & \checkmark & & 80.46 & 80.95 & 86.98 \\

\midrule
\midrule

LAA-Net~\cite{laa_net} & CVPR'24 & \checkmark & $\times$ & & 86.28 & 81.12 & 90.15 \\

LAA-Former-S & - & \checkmark & $\times$ & & 88.23 & \textbf{91.58} & 90.99 \\
LAA-Former-B & - & \checkmark & $\times$ & & \textbf{90.93} & \underline{90.52} & 91.29 \\

LAA-Swin-S & - & \checkmark & $\times$ & & 88.30 & 90.31 & \underline{92.57} \\
LAA-Swin-B & - & \checkmark & $\times$ & & \underline{89.39} & 89.81 & \textbf{93.73} \\

\bottomrule
\end{tabular}%
}}
\vspace{-2mm}
\label{tabl:diff_eval}
\end{table}

\begin{table}
\centering
\caption{\textbf{Comparison in terms of numbers of parameters (\#Para.) and AUC (\%) at the \textit{video-level}} using cross-manipulation evaluation on five subsets of DF40~\cite{df40}. For the sake of clarity, we note that we report the (\#Para.) for the entire model, including all auxiliary branches. These branches can be removed at the inference for more efficient computation as discussed in Section~\ref{sec:laa-x}.
}
\resizebox{\linewidth}{!}{
{\rowcolors{7}{Plum!10}{SkyBlue!20}
\begin{tabular}{c c c cc ccc}
\toprule
\multirow{2}{*}{Method} & \multirow{2}{*}{Venue} & \multirow{2}{*}{\#Para.} & \multicolumn{5}{c}{Cross-manipulation} \\
\cmidrule(lr){4-8}
& & & E4S & FOMM & BlendFace & FSGAN & MobileSwap \\
\midrule
\midrule
SBI$^\ast$~\cite{sbi} & CVPR'22 & 19M & 52.80 & 79.56 & 86.50 & 85.36 & 86.64 \\
FAViT$^\ast$~\cite{FAViT} & TCSVT'24 & 128M & 74.70 & 76.99 & 88.43 & \underline{96.96} & 83.96 \\
StA+VB~\cite{plug_play} & CVPR'25 & 353M & - & - & 90.6 & 96.4 & 94.6 \\

\midrule
\midrule

LAA-Net~\cite{laa_net} & CVPR'24 & 27M & 81.70 & \textbf{88.29} & 91.28 & \textbf{97.52} & \underline{97.15} \\
LAA-Former-S & - & 23M & 88.89 & 82.34 & 91.07 & 95.45 & 93.40 \\
LAA-Former-B & - & 91M & 86.19 & \underline{84.11} & \underline{93.23} & 94.18 & 95.92 \\

LAA-Swin-S & - & 55M & \underline{90.79} & 80.43 & \textbf{93.52} & 94.98 & \textbf{97.87} \\
LAA-Swin-B & - & 91M & \textbf{91.42} & 81.82 & 91.77 & 95.77 & 96.96 \\

\bottomrule
\end{tabular}%
}}
\vspace{-3mm}
\label{tabl:df40_eval}
\end{table}

\vspace{1mm}
\noindent\textbf{Evaluation Metrics.}
\label{subsec:eval_metrics}
To compare the performance of LAA-X with SOTA methods, we report the common Area Under the Curve (AUC) and the Average Precision (AP) as in~\cite{fxray, cstency_learning, sbi, caddm}. 
More metrics, namely, Average Recall (AR), and mean F1-score (mF1), are provided in supplementary materials.

\vspace{1mm}
\noindent\textbf{Data Pre-processing.}
\label{subsec:preprocessing}
Following the splitting convention of~\cite{ff++}, we extract $128$, $32$, and $32$ frames from each video for training, validation, and testing, respectively.
RetinaNet~\cite{retina_face} is used to crop faces with a conservative enlargement (by a factor of $1.25$) around the face center. Note that all the cropped images are then resized to $384 \times 384$ for LAA-Net, $112 \times 112$ for LAA-Former-S, and $224 \times 224$ for LAA-Former-B. 
In addition, we utilize Dlib~\cite {dlib} to extract and store $68$ and $81$ facial landmarks for each frame.
Finally, the preserved landmark keypoints are leveraged to dynamically generate pseudo-fakes during each training iteration.

\vspace{1mm}
\noindent\textbf{Implementation Details.} We applied different training strategies to the two versions of LAA-X. 
For \textit{1) LAA-Net: }
it is trained for $100$ epochs with the SAM optimizer~\cite{sam}, a weight decay of $10^{-4}$, and a batch size of $16$. We apply a learning rate scheduler that increases from $5. 10^{-5}$ to $2. 10^{-4}$ in the first quarter of the training and then decays to zero in the remaining quarters. The backbone is initialized with pretrained weights on ImageNet~\cite{imagenet}. During training, we freeze the backbone for a warm-up at the first $6$ epochs and only train the remaining layers. The parameters $\lambda_1$ and $\lambda_2$, defined in Eq.~\eqref{equa:total_loss}, are set to $10$ and $100$, respectively. All experiments are carried out using a GPU Tesla V-$100$.
Regarding \textit{2) LAA-Former: }
We train the model for $200$ epochs using the AdamW~\cite{adamw} optimizer with a weight decay of $10^{-4}$ and a batch size of $32$. The weights are initialized using pretrained DINO~\cite{dinov1} on ImageNet~\cite{imagenet}. The learning rate is maintained at $5 \times 10^{-5}$ during the first quarter of iterations, then gradually decays to zero over the remaining epochs. We freeze the backbone (i.e., ViT without the head) for the first $6$ epochs, before training all layers. 
The $\lambda_{att}$ in Eq.~(\ref{equa:total_loss_transformer}) is empirically set to $10$.
All experiments are conducted on $4$ NVIDIA A100 GPUs.

For data augmentation, we apply horizontal flipping, random cropping, random scaling, random erasing~\cite{random_erasing}, color jittering, Gaussian noise, blurring, and JPEG compression. Furthermore, label smoothing~\cite{label_smoothing} is utilized and integrated into the loss function as a regularizer. To generate pseudo-fakes, two blending synthesis techniques are considered, namely, Blended Images (BI)~\cite{fxray} and Self-Blended Images (SBI)~\cite{sbi}. 
During training, in each epoch, for each video in the batch data, we dynamically randomize only $k$ frames, with $k=8$ or $k=16$ when using SBI~\cite{sbi} or BI~\cite{fxray}, respectively.

\vspace{1mm}
\noindent\textbf{Architecture Choices.} 
We adopt the B$4$ variant (EFN-B4) of the EfficientNet~\cite{efn_net} as the backbone for LAA-Net. Regarding LAA-Former, we employ two variants that we call LAA-Former-S and LAA-Former-B. \textit{By default}, we use the lightweight LAA-Former-S where $H=W=112$ and $P=8$, while utilizing $H=W=224$ and $P=16$ for LAA-Former-B. LAA-Former is based on a vanilla vision transformer, i.e., ViT~\cite{ViT}. Although LAA-Former is based on ViT, we also assess its applicability using another transformer architecture, namely \textit{LAA-Swin}, which is based on Swin~\cite{swin}. Similarly to LAA-Former, we consider two variants: LAA-Swin-S and LAA-Swin-B. Additional architectural details are provided in the supplementary materials.

\subsection{Comparison with State-of-the-art Methods}

\noindent\textbf{Generalization to Unseen Datasets.}
To assess the generalization capabilities of our method, we evaluate LAA-X under the challenging cross-dataset setup ~\cite{ete_recons, sfdg, fakestormer, FAViT, LSDA, aunet, caddm, sbi, diffusionfake}. Table~\ref{tabl:cross_auc_full_metrics_2} and Table~\ref{tabl:diff_eval} report the results obtained on multiple unseen datasets, i.e., CDF2~\cite{celeb_df}, DFW~\cite{wdf}, DFD~\cite{dfd}, DFDCP~\cite{dfdcp}, DFDC~\cite{dfdc}, and DiffSwap~\cite{diffusionface} at the video-level and the frame-level, respectively. 

It can be observed that LAA-X achieves state-of-the-art results on most considered benchmarks, especially on the large-scale DFDC dataset, the unknown in-the-wild DFW, and the recent diffusion-based DiffSwap.
Although LAA-X builds on the blending assumption, this suggests that explicitly focusing on vulnerabilities rather than directly estimating blending masks allows better detecting non-blending-based face-swaps,
demonstrating its generalizability and robustness to different qualities of deepfakes. 
Particularly, LAA-Net clearly outperforms other attention-based approaches such as Multi-attentional~\cite{multi-attentional} and SFDG~\cite{sfdg} by a considerable margin of $27.14$\% and $19.57$\% in terms of AUC and AP on CDF2, respectively. 
The best performance is reached when using SBI as a data synthesis, confirming the importance of modeling generic and subtle artifacts. 
An exception is that the performance of LAA-Net (w/ BI) is slightly superior to LAA-Net (w/ SBI) only on DFD, with an improvement of $1.08$\%  and $0.4$\% of AUC and AP, respectively. 
A plausible explanation could be the fact that deepfake artifacts in DFD are less challenging to detect or possibly similar to those in FF++. In fact, numerous methods report AUC and AP scores exceeding $98$\%.

Furthermore, despite its simplicity, the compromise between the implicit SA and the explicit local attention (L2-Att) to artifact-prone vulnerable patches allows LAA-Former to improve the average performance by $2.85 \%$ (w/ SBI) and $5.6 \%$ (w/ BI) as compared to LAA-Net in Table~\ref{tabl:cross_auc_full_metrics_2}. This suggests the importance of modeling both local features and global semantics modeling. It is also noted that scaling the model leads to a decent increase in overall performance (Table~\ref{tabl:diff_eval}).

\vspace{1mm}
\noindent\textbf{In-dataset Evaluation.} 
We compare the performance of LAA-X to existing methods under the in-dataset protocol as in~\cite{cstency_learning, caddm, sbi, aunet, sfdg, altfreezing}. The first column in Table~\ref{tabl:cross_auc_full_metrics_2} reports the obtained results on the testing set of FF++. It can be seen that all methods achieve competitive performance on the forgeries of the FF++ dataset. Our method, combined with SBI, outperforms all methods with an AUC of $99.96$\%, while using only real data for training. 

Furthermore, we report in Table~\ref{tabl:df40_eval} the generalization performance of LAA-Net and several variants of LAA-Former under a cross-manipulation evaluation setting on five subsets of the recent large-scale DF40~\cite{df40} dataset. Our method achieves notably higher AUC scores than other methods across all subsets, highlighting its strong generalization capability under diverse ranges of unseen manipulation techniques.

\begin{table}[!t]
\centering
\caption{\textbf{Robustness to unseen perturbations.}}
\resizebox{\linewidth}{!}{
{\rowcolors{11}{Plum!10}{SkyBlue!20}
\begin{tabular}{c cc ccccHcHc}
\toprule
\multirow{2}{*}{Method} & \multicolumn{2}{c}{Training set} & \multicolumn{8}{c}{Perturbation set} \\
\cmidrule(lr){2-3}
\cmidrule(lr){4-10}
 & Real & Fake & Saturation & Contrast & Block & Noise & Blur & Pixel & V Compression & Avg. \\ 
\midrule
\midrule
Xception~\cite{xception} & {$\checkmark$} & \checkmark & 99.3 & 98.6 & 99.7 & 53.8 & 60.2 & 74.2 & 62.1 & 85.12 \\

FaceXray~\cite{fxray} & {$\checkmark$} & \checkmark & 97.6 & 88.5 & 99.1 & 49.8 & 63.8 & 88.6 & 55.2 & 84.72 \\

CNN-aug~\cite{CNN_AUG} & {$\checkmark$} & \checkmark & 99.3 & 99.1 & 95.2 & 54.7 & 76.5 & 91.2 & 72.5 & 87.90 \\

LipForensics~\cite{lipforensics} & {$\checkmark$} & \checkmark & \underline{99.9} & 99.6 & 87.4 & 73.8 & 96.1 & \underline{95.6} & 95.6 & 91.26 \\

SBI~\cite{sbi} & $\checkmark$ & $\times$ & 92.0 & 92.3 & 92.2 & 62.2 & 72.7 & 79.1 & 79.1 & 83.56 \\

LSDA~\cite{LSDA} & $\checkmark$ & $\checkmark$ & 98.7 & 94.4 & 98.3 & 66.4 & - & 90.7 & 90.7 & 89.70 \\

\midrule
\midrule

LAA-Net~\cite{laa_net} & \checkmark & $\times$ & \textbf{99.96} & \textbf{99.96} & \textbf{99.96} & 53.90 & \underline{98.22} & \textbf{99.80} & - & 90.72 \\ 

LAA-Former & \checkmark & $\times$ & 98.04 & 95.96 & 97.02 & \underline{75.28} & 87.38 & 91.14 & 91.14 & \underline{91.49} \\ 

LAA-Swin & \checkmark & $\times$ & 99.79 & \underline{99.77} & \underline{99.92} & \textbf{81.60} & 89.80 & 94.88 & - & \textbf{95.19} \\ 

\bottomrule
\end{tabular}%
}}
\vspace{-2mm}
\label{tabl:ff_noise_auc}
\end{table}

\vspace{1mm}
\noindent\textbf{Robustness to Unseen Perturbations.}\label{noise_robust} Since deepfakes can be easily spread and altered on various social platforms, the robustness of LAA-X against some unseen common perturbations is investigated. Following the settings of~\cite{DFo}, we evaluate the robustness of LAA-X across five unseen corruptions. The results are reported in Table~\ref{tabl:ff_noise_auc}, using models trained on FF++. 
As LAA-Net focuses on vulnerable points, it can be seen that color-related changes, such as Saturation and Contrast, do not impact the performance. 
However, the proposed method is extremely sensitive to structural perturbations such as Gaussian Noise. 
One possible reason is due to the introduction of noise that elevates the difficulty of detecting the vulnerable points. To confirm that, we report in the supplementary materials the inference output of the heatmap before and after applying a Gaussian Noise on a facial image.

On the other side, thanks to vulnerable patches, LAA-Former and LAA-Swin show substantially improved robustness to noise. Although they are slightly more affected by some distortions than LAA-Net, both transformer-based architectures do improve the overall performance.

\begin{figure}[t]
    \centering
    \includegraphics[width=\linewidth]{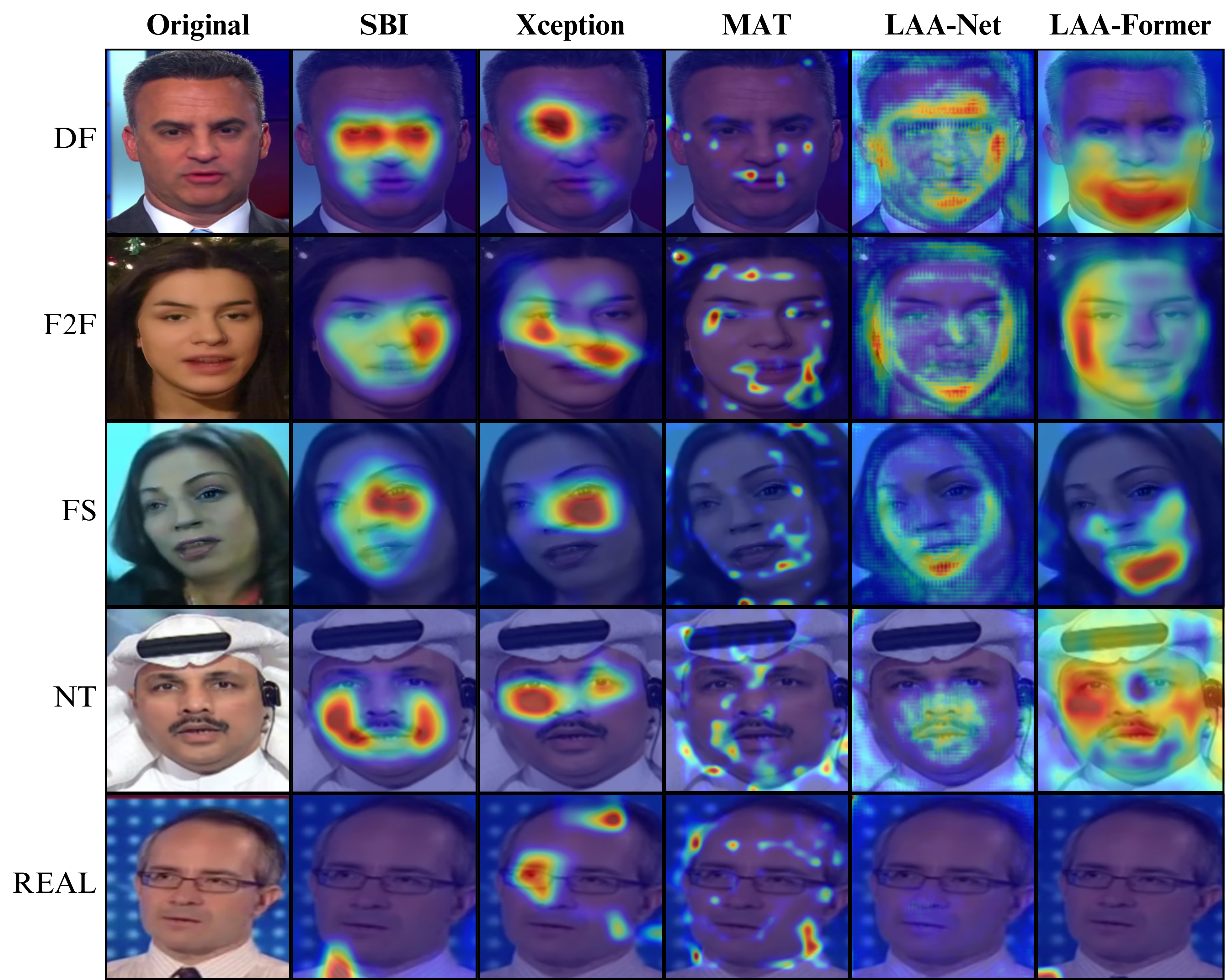}
    \caption{\textbf{Visualization of saliency maps} on different types of manipulation from FF++~\cite{ff++}. LAA-Net and LAA-Former is compared to SBI~\cite{sbi}, Xception~\cite{ff++}, and MAT~\cite{multi-attentional}.}
\label{fig:qualitative_results}
\vspace{-3mm}
\end{figure}

\vspace{1mm}
\noindent\textbf{Qualitative Results.}\label{abl:qua_res}
We provide Grad-CAMs~\cite{gradCAM} in Figure~\ref{fig:qualitative_results}, to visualize the image regions in deepfakes that are activated by LAA-Net, LAA-Former, SBI~\cite{sbi}, Xception~\cite{ff++}, and Multi-attentional (MAT)~\cite{multi-attentional} on FF++~\cite{ff++}. Generally, attention-based methods such as MAT~\cite{multi-attentional}, LAA-Net, and LAA-Former focus more on localized regions. However, in some cases, MAT~\cite{multi-attentional} concentrates on irrelevant regions such as the background or the inner face areas, even on real data. Conversely, LAA-X consistently identifies blending artifacts and shows interesting capabilities on expression-manipulated Neural Textures (NT).

\begin{table*}[ht]
\caption{\textbf{Traditional FPN versus E-FPN} using the SBI data synthesis under the cross-dataset evaluation protocol.
We report the results when integrating features $\mathbf F^{(i)}$ from different layers.}
\centering
\resizebox{\linewidth}{!}{
\begin{tabular}{c|c|cccc|HHcc|cc|cc|cc}
\hline
& \multicolumn{5}{c|}{EFN-B4} & \multicolumn{10}{c}{Test Set AUC (\%)} \\
\hline
\multicolumn{2}{c|}{} & \multicolumn{4}{c|}{E-FPN Integration} & \multicolumn{2}{H}{CDF1} & \multicolumn{2}{c|}{CDF2} & \multicolumn{2}{c|}{DFD} & \multicolumn{2}{c|}{DFW} & \multicolumn{2}{c}{DFDCP} \\
\hline
& {$\mathbf F^{(6)}$} &$\mathbf F^{(5)}$ & $\mathbf F^{(4)}$ & $\mathbf F^{(3)}$ & {$\mathbf F^{(2)}$} & FPN & {E-FPN} & FPN & {E-FPN} & FPN & {E-FPN} & FPN & {E-FPN} & FPN & E-FPN \\
\hline
\hline
(a) & \checkmark & & & & {} & \multicolumn{2}{H}{92.72} & \multicolumn{2}{c|}{91.56} & \multicolumn{2}{c|}{98.27} & \multicolumn{2}{c|}{73.02} & \multicolumn{2}{c}{78.35} \\

(b) & \checkmark & \checkmark & & & {} & 93.37 & {{93.81}} & {93.42} & {91.79} & {98.59} & {97.12} & {73.78} & {71.39} & {78.40} & 75.80 \\

(c) & \checkmark & \checkmark & \checkmark & & {} & 89.45 & {\underline{{95.34}}} & 88.72 & {{92.86}} & 97.96 & {\textbf{{98.95}}} & 69.40 & {\underline{{74.93}}} & 71.91 & \underline{83.97} \\

(d) & \checkmark & \checkmark & \checkmark & \checkmark & {} & 88.97 & {{93.11}} & 88.35 & {\textbf{{95.40}}} & \underline{98.89} & {98.43} & 70.94 & {\textbf{{80.03}}} & 79.02 & \textbf{{86.94}} \\

(e) & \checkmark & \checkmark & \checkmark & \checkmark & \checkmark & 92.59 & {\textbf{{95.64}}} & 92.16 & {\underline{{94.22}}} & 96.58 & {97.31}& 65.17 & {{72.54}} & 74.31 & {82.90} \\

\hline
\hline

\multicolumn{6}{c|}{Avg.} &  & {\textbf{{-}}} & 90.84 & \textit{93.16 \textcolor{ForestGreen}{$\uparrow$(2.32)} } & \textit{98.06} & 98.02 \textcolor{Salmon}{$\downarrow$(0.04)} & 70.46 & \textit{74.38 \textcolor{ForestGreen}{$\uparrow$(3.92)}} & 76.40 & \textit{81.59 \textcolor{ForestGreen}{$\uparrow$(5.19)}} \\

\hline
\end{tabular}
}
\vspace{-3mm}
\label{tabl:efpn_tfpn}
\end{table*}

\begin{figure}[t]
    \centering
    \includegraphics[width=\linewidth]{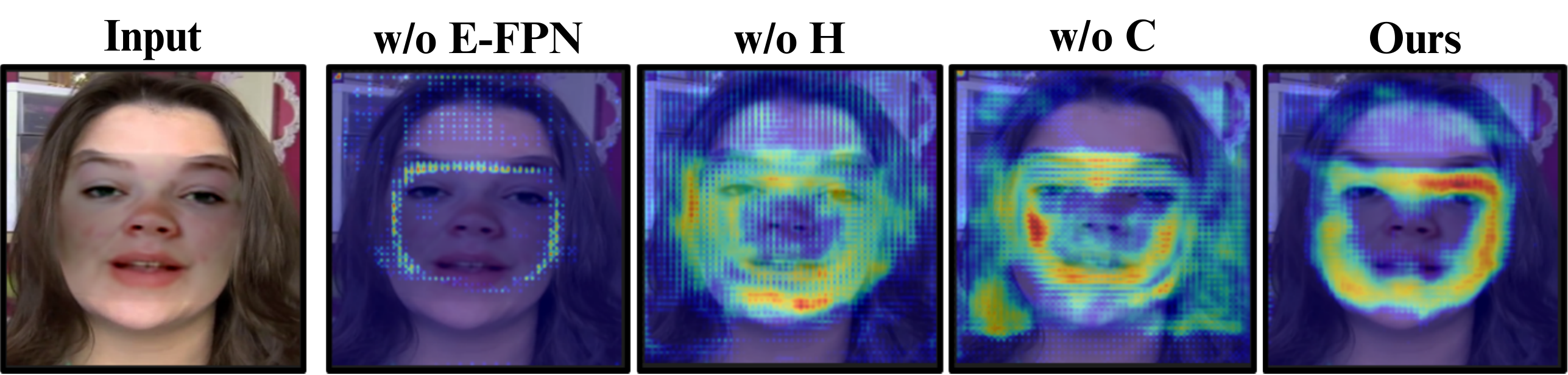}
    \caption{\textbf{Visualization of saliency maps of different components in LAA-Net} w/o E-FPN, w/o H, and w/o C refer to ablating E-FPN, heatmap branch, and self-consistency branch, respectively.}
    \label{fig:efpn_com_abl}
    \vspace{-2mm}
\end{figure}

\begin{table}
\centering
\centering
\captionof{table}{\textbf{Ablation study of LAA-Net's components} including the Consistency branch (C), Heatmap branch (H), and E-FPN. \label{tabl:laa_components_eval}}
\resizebox{\linewidth}{!}{
\begin{tabular}{ccc Hcccc c}
\toprule
\multirow{2}{*}{C} & \multirow{2}{*}{H} & \multirow{2}{*}{E-FPN} & \multicolumn{5}{c}{Test set AUC (\%)} \\ 
 \cmidrule{4-8} 
 &  &  & {CDF1} & CDF2 & DFD & DFDCP & DFW & Avg. \\ 
 
\midrule
\midrule

\textcolor{gray}{$\times$} & \textcolor{gray}{$\times$} & \textcolor{gray}{$\times$} & - & 74.54 & 92.24 & 70.81 & 59.81 & 74.35 \\

\textcolor{gray}{$\times$} & $\checkmark$ & \textcolor{gray}{$\times$} & - & 80.89 & 94.53 & 77.93 & 67.12 & 80.12 \textcolor{ForestGreen}{$\uparrow$(5.77)} \\

\textcolor{gray}{$\times$} & \textcolor{gray}{$\times$} & $\checkmark$ & - & 84.21 & 95.03 & 80.68 & 65.47 & 81.35 \textcolor{ForestGreen}{$\uparrow$(7.00)} \\

\textcolor{gray}{$\times$} & $\checkmark$ & $\checkmark$ & \textbf{97.82} & \textbf{95.56} & \textbf{98.54} & \underline{82.21} & \underline{74.98} & \underline{87.82} \textcolor{ForestGreen}{$\uparrow$(13.47)} \\

$\checkmark$ & \textcolor{gray}{$\times$} & $\checkmark$ & 84.13 & 79.87 & 94.60 & 71.70 & 72.47 & 79.66 \textcolor{ForestGreen}{$\uparrow$(5.31)} \\

$\checkmark$ & $\checkmark$ & \textcolor{gray}{$\times$} & 92.72 & 91.56 & 98.27 & 78.35 & 73.02 & 85.30 \textcolor{ForestGreen}{$\uparrow$(10.95)} \\

$\checkmark$ & $\checkmark$ & $\checkmark$ & \underline{93.11} & \underline{95.40} & \underline{98.43} & \textbf{86.94} & \textbf{80.03} & \textbf{90.20 \textcolor{ForestGreen}{$\uparrow$(15.85)}} \\

\bottomrule
\end{tabular}
}
\vspace{-3mm}
\end{table}

\subsection{E-FPN versus Traditional FPN} 
To assess the effectiveness of the low-level features injected by E-FPN into the final feature representation, we combine different feature levels and compare the results of E-FPN and traditional FPN~\cite{fpn_obdet, focal_loss} in Table~\ref{tabl:efpn_tfpn}. It can be seen that in general E-FPN outperforms FPN except for $\mathbf{F}^{(5)}$. This confirms the relevance of employing multi-scale features and the need for reducing their redundancy in the context of deepfake detection.

\subsection{Additional Discussions on CNN-based LAA-X: LAA-Net}

\noindent\textbf{Ablation Study of the LAA-Net's Components. } Table~\ref{tabl:laa_components_eval} reports the cross-dataset performance of LAA-Net when discarding the following components:  E-FPN, the consistency branch denoted by C and the heatmap branch denoted by H. The best performance is reached when all the components are integrated. It can be seen that the proposed explicit attention mechanism through the heatmap branch contributes more to improving the result. A qualitative example visualizing Grad-CAMs~\cite{gradCAM} with different components of LAA-Net is also given in Figure~\ref{fig:efpn_com_abl}. The illustration clearly shows that by combining the three components, the network activates more precisely the blending region.

\begin{table}[ht]
\caption{\textbf{Sensitivity analysis.} The impact of the hyperparameters $\lambda_1$ and $\lambda_2$ using the cross-dataset protocol on three datasets in terms of AUC.}
\centering
\begin{tabular}{cc Hccc c} 
\toprule
 \multirow{2}{*}{$\lambda_1$} & \multirow{2}{*}{$\lambda_2$} & \multicolumn{5}{c}{Test Set AUC (\%)} \\
 \cmidrule{3-6}
                                                      & & {CDF1} & CDF2 & DFDCP & {DFW} & Avg.           \\ 
\midrule
\midrule

1 & 1 & {94.20} & 90.69 & 78.12 & {70.98} & 79.93 \\
10 & 10 & {\underline{96.60}} & \textbf{95.73} & \underline{85.87} & 73.56 & \underline{85.05} \\
100 & 100 & {\textbf{97.30}} & 93.72 & 78.60 & {75.25} & 82.52 \\
100 & 10 & {93.29} & 93.05 & 83.86 & {\underline{76.72}} & 84.54 \\
10 & 100 & {93.11} & \underline{95.40} & \textbf{86.94} & {\textbf{80.03}} & \textbf{87.46} \\

\bottomrule
\end{tabular}
\vspace{-2mm}
\label{tabl:blance_factor_eval}
\end{table}

\vspace{1mm}
\noindent\textbf{Sensitivity Analysis.}
In this subsection, we analyze the impact of the two hyperparameters $\lambda_1$ and $\lambda_2$ given in Eq.~\eqref{equa:total_loss}. Table~\ref{tabl:blance_factor_eval} shows the experimental results for different values of $\lambda_1$ and $\lambda_2$. It can be noted that our model is robust to different hyperparameter values, with the best average performance obtained with $\lambda_1=10$ and $\lambda_2=100$.

\begin{figure}[!h]
    \centering
    \includegraphics[width=\linewidth]{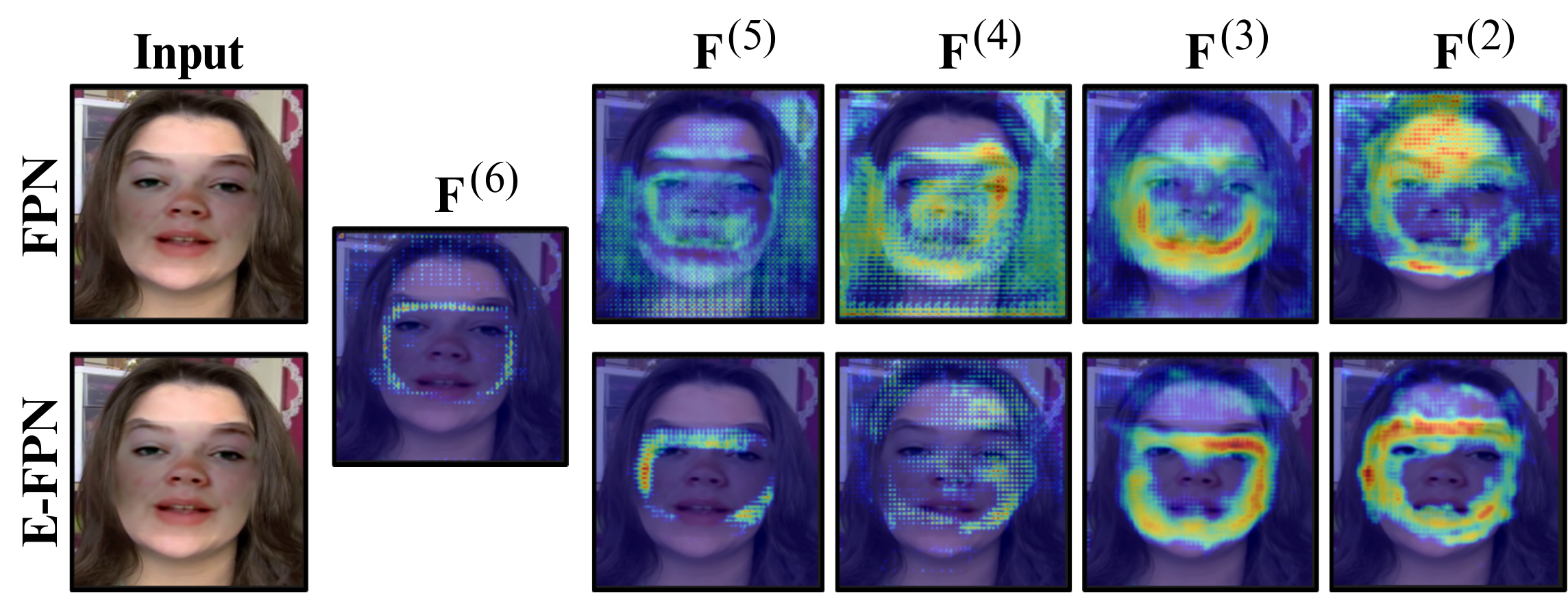}
    \caption{\textbf{Visualization of saliency maps} using \textit{E-FPN} and \textit{FPN} with different integration of multi-scale layers. We can see that E-FPN can focus better on artifacts as compared to FPN. The setup details are provided in Table~\ref{tabl:efpn_tfpn}.}
    \label{fig:efpn_fpn_abl}
    \vspace{-3mm}
\end{figure}

\vspace{1mm}
\noindent\textbf{Qualitative Results: E-FPN versus FPN. }
A qualitative comparison between the proposed  E-FPN  and the traditional FPN with different fusion settings is reported in Figure~\ref{fig:efpn_fpn_abl}. Using EFN-B4 as our backbone, the $\mathbf F^{(6)}$ refers to the features extracted from the last convolution block in the backbone. In other words, this means that no FPN design is integrated. By gradually aggregating features from lower to higher resolution layers, we can observe the improvement of the forgery localization ability for both E-FPN and FPN. More notably, E-FPN produces more precise activations on the blending boundaries as compared to FPN. This can be explained by the fact that the E-FPN integrates a filtering mechanism for learning less noise. In contrast, FPN seems to consider regions outside the blending boundary, which results in lower performance as previously shown in Table~\ref{tabl:efpn_tfpn}.

\subsection{Additional Discussions on Transformer-based LAA-X: LAA-Former}
\noindent\textbf{Effect of Patch Size.}\label{abl:patchsize_effects}
As shown in Figure~\ref{fig:ViT_CNN_losses}, patch size has a clear effect on ViT’s learning capability under different training configurations. In this section, we further investigate the impact of patch size on generalization performance by varying the input resolution and patch size of LAA-Former trained on FF++~\cite{ff++} and tested on unseen deepfakes. Table~\ref{tabl:patch_size_abl} presents the cross-dataset evaluation results across several datasets~\cite{celeb_df, dfd, wdf, dfdcp, dfdc}. We observe that either reducing the patch size or increasing the input resolution improves model performance, confirming the importance of patch size in transformer-based architectures for extracting localized features in deepfake detection.

\begin{table}
\centering
\caption{\textbf{Effect of input resolution and patch size.} 
}
\resizebox{\linewidth}{!}{
\begin{tabular}{c c Hccc c c c}
\toprule

\multirow{2}{*}{Res. \& Pat.} & \multicolumn{7}{c}{Test set AUC (\%)} \\
\cmidrule{2-8}
 & FF++ & CDF1 & CDF2 & DFD & DFW & DFDCP & DFDC & Avg. \\
 
\midrule
\midrule

$112$P$16$ & 81.83 & 87.93 & 85.26 & 73.56 & 76.87 & 93.28 & 72.53 & 80.56 \\
$112$P$8$ & 97.67 & 97.25 & 94.45 & 96.12 & 81.74 & 96.30 & 78.91 & 90.87 \\
$224$P$8$ & \textbf{99.93} & \textbf{99.00} & \textbf{96.84} & \textbf{99.54} & \textbf{82.11} & \textbf{96.99} & \textbf{79.01} & \textbf{92.40} \\

\bottomrule
\end{tabular}%
}
\vspace{-2mm}
\label{tabl:patch_size_abl}
\end{table}

\begin{table}[]
\centering
\caption{\textbf{Ablation study of LAA-Former's components.}}
\resizebox{\linewidth}{!}{
\begin{tabular}{c c c Hc HH Hc Hc Hc c}
\toprule
\multirow{2}{*}{Model} & \multirow{2}{*}{Lr} & \multirow{2}{*}{L2-Att} & \multicolumn{11}{c}{Test set AUC (\%)} \\
\cmidrule{5-13}
& & & \multicolumn{2}{c}{FF++} & \multicolumn{2}{H}{CDF1} & \multicolumn{2}{c}{CDF2} & \multicolumn{2}{c}{DFD} & \multicolumn{2}{c}{DFDC} & Avg. \\

\midrule
\midrule

ViT~\cite{ViT} (w/ SBI) & \multirow{2}{*}{$1 \times 10^{-3}$} & \textcolor{gray}{$\times$} & - & 68.86 
& - & - 
& - & 67.29 
& - & 60.58 
& - & 61.32 
& 64.51 \\ 
LAA-Former & & $\checkmark$ & - & 73.52 
 & - & - 
 & - & 71.78 
 & - & 65.34 
 & - & 62.51 
 & 68.29 (\textcolor{ForestGreen}{$\uparrow$3.78}) \\ 
 
\midrule

ViT~\cite{ViT} (w/ SBI) & \multirow{2}{*}{$5 \times 10^{-4}$} & \textcolor{gray}{$\times$} & - & 75.68 
& - & - 
& - & 72.99 
& - & 57.19
& - & 62.38
& 67.06 \\ 
LAA-Former & & $\checkmark$ & - & 80.08 
 & - & - 
 & - & 87.45 
 & - & 65.62 
 & - & 71.04 
 & 76.05 (\bf{\textcolor{ForestGreen}{$\uparrow$8.99}}) \\ 
 
\midrule

ViT~\cite{ViT} (w/ SBI) & \multirow{2}{*}{$1 \times 10^{-4}$} & \textcolor{gray}{$\times$} & - & 95.99 
& - & - 
& - & 89.27 
& - & 89.71 
& - & 78.50 
& 88.36 \\ 
LAA-Former & & $\checkmark$ & - & 96.14 
 & - & - 
 & - & \bf{95.20} 
 & - & 91.14 
 & - & 78.85 
 & 90.33 (\textcolor{ForestGreen}{$\uparrow$1.97}) \\ 
 
\midrule

ViT~\cite{ViT} (w/ SBI) & \multirow{2}{*}{$5 \times 10^{-5}$} & \textcolor{gray}{$\times$} & - & 97.48 
& 85.18 & 94.42 
& 83.96 & 92.62 
& 92.71 & 95.72 
& 77.65 & 77.35 
& 90.79 \\ 
LAA-Former & & $\checkmark$ & - & \textbf{97.67} 
 & \textbf{89.53} & \textbf{97.25} 
 & \textbf{90.34} & 94.45
 & \textbf{93.42} & \textbf{96.12} 
 & \textbf{78.71} & \textbf{78.91} 
 & \textbf{91.79} (\textcolor{ForestGreen}{$\uparrow$1.00}) \\ 

\midrule
\midrule

Swin~\cite{swin} (w/ SBI) & \multirow{2}{*}{$1 \times 10^{-3}$} & \textcolor{gray}{$\times$} & - & 99.48 
& - & - 
& - & 81.37 
& - & 96.05 
& - & 66.69 
& 85.89 \\ 
LAA-Swin & & $\checkmark$ & - & 99.88 
 & - & - 
 & - & 94.91 
 & - & 97.17 
 & - & 74.33 
 & 91.57 (\textcolor{ForestGreen}{$\uparrow$\bf{5.68}}) \\ 

\midrule

Swin~\cite{swin} (w/ SBI) & \multirow{2}{*}{$5 \times 10^{-4}$} & \textcolor{gray}{$\times$} & - & \bf{99.98} 
& - & - 
& - & 89.00 
& - & 98.94 
& - & 71.15 
& 89.76 \\ 
LAA-Swin & & $\checkmark$ & - & 99.97
 & - & -
 & - & \bf{95.43 }
 & - & 99.58 
 & - & 74.97 
 & 92.49 (\textcolor{ForestGreen}{$\uparrow$2.73}) \\ 

\midrule

Swin~\cite{swin} (w/ SBI) & \multirow{2}{*}{$1 \times 10^{-4}$} & \textcolor{gray}{$\times$} & - & 99.89 
& - & - 
& - & 90.18 
& - & 99.54 
& - & 73.38 
& 90.74 \\ 
LAA-Swin & & $\checkmark$ & - & \bf{99.98} 
 & - & - 
 & - & 93.87 
 & - & 99.62 
 & - & \bf{77.92} 
 & 92.85 (\textcolor{ForestGreen}{$\uparrow$2.11}) \\ 

\midrule

Swin~\cite{swin} (w/ SBI) & \multirow{2}{*}{$5 \times 10^{-5}$} & \textcolor{gray}{$\times$} & - & 99.75 
& 71.23 & 94.46 
& 82.50 & 90.89 
& 99.89 & 99.59 
& \textbf{69.94} & 74.20 
& 91.11 \\ 
LAA-Swin & & $\checkmark$ & - & 99.89
 & \textbf{75.85} & \textbf{96.69} 
 & \textbf{83.29} & 94.48
 & \textbf{99.94} & \textbf{99.68} 
 & 68.48 & 77.47 
 & \textbf{92.88} (\textcolor{ForestGreen}{$\uparrow$1.77}) \\ 

\bottomrule
\end{tabular}%
}
\label{tabl:lga_abl}
\vspace{-2mm}
\end{table}

\vspace{1mm}
\noindent\textbf{Ablation study of LAA-Former's components.}\label{abl:l2-att_effects}
The plug-and-play L2-Att plays a crucial role in explicitly guiding ViT~\cite{ViT}/Swin~\cite{swin} to attend to artifact-prone vulnerable patches. To validate its impact in our proposed architectures, we compare the baseline models (w/o L2-Att) with LAA-Former (ViT+L2-Att)/LAA-Swin (Swin+L2-Att). All models are trained with SBI~\cite{sbi}. The results on several datasets~\cite{ff++, celeb_df, dfd, dfdc} are presented in Table~\ref{tabl:lga_abl}. 
As shown, L2-Att consistently contributes to the enhancement of both ViT and Swin, confirming the relevance of the proposed explicit attention mechanism.

\begin{table}[]
\centering
\caption{\textbf{Vulnerable Patches (V-Patch) vs. Vulnerable Points (V-Point).} 
}
\scalebox{0.8}{
\begin{tabular}{c c c c HHccc c}
\toprule
\multirow{2}{*}{Model} & \multirow{2}{*}{Target} & \multirow{2}{*}{\#Para.} & \multirow{2}{*}{FLOPs} & \multicolumn{5}{c}{Test set AUC(\%)}\\
\cmidrule{5-9}
 & & & & FF++ & CDF1 & CDF2 & DFD & DFDC & Avg. \\

\hline
\hline
\multirow{2}{*}{LAA-Former} & V-Point & 23.61M & 9.4G & - & 97.22 & 93.39 & 93.70 & 77.71 
& 88.26 \\ 
 & V-Patch & 22.77M & 8.9G & - & \textbf{97.25} & \textbf{94.45} & \textbf{96.12} & \textbf{78.91} 
 & \textbf{89.83(\textcolor{ForestGreen}{$\uparrow$1.67})} \\ 

\midrule

\multirow{2}{*}{LAA-Swin} & V-Point & 57.25M & 8.1G & - & 96.16 & 94.25 & 99.30 & 74.99 
& 89.51 \\ 
 & V-Patch & 54.89M & 6.5G & - & \textbf{96.69} & \textbf{94.48} & \textbf{99.68} & \textbf{77.47} 
 & \textbf{90.54(\textcolor{ForestGreen}{$\uparrow$1.03})} \\ 

\bottomrule
\end{tabular}%
}
\label{tabl:vpart_vpoint_abl}
\vspace{-2mm}
\end{table}

\begin{table}
\centering
\caption{\textbf{Selection of $f_2$}. 
}
\begin{tabular}{c H Hccc H c c}
\toprule
\multirow{2}{*}{$f_2$} & \multicolumn{7}{c}{Test set AUC (\%)} \\
\cmidrule(lr){2-8}
 & FF++ & CDF1 & CDF2 & DFD & DFW & DFDCP & DFDC & Avg. \\
 
\midrule
\midrule

mean & 97.45 & 96.29 & 94.16 & 95.03 & 81.02 & - & 78.28 & 87.12  \\
max & \textbf{97.67} & \textbf{97.25} & \textbf{94.45} & \textbf{96.12} & \textbf{81.74} & \textbf{96.30} & \textbf{78.91} & \textbf{87.81(\textcolor{ForestGreen}{$\uparrow$0.69})} \\

\bottomrule
\end{tabular}%
\label{tabl:vul_type_abl}
\vspace{-2mm}
\end{table}

\begin{table}
\centering
\caption{\textbf{Impact of loss balancing factor $\lambda_{att}$ (Eq.~\eqref{equa:total_loss_transformer}.} 
}
\begin{tabular}{c H H c c c c}
\toprule
\multirow{2}{*}{$\lambda_{att}$} & \multicolumn{5}{c}{Test set AUC (\%)} \\
\cmidrule{2-6}
 & FF++ & CDF1 & CDF2 & DFD & DFDC & Avg. \\ 
\midrule
\midrule

1 & - & 94.68 & 93.67 & 94.62 & 77.91 
& 88.73 \\ 

10 & - & 97.25 & \underline{94.45} & \textbf{96.12} & \textbf{78.91} 
& \textbf{89.83} \\ 

100 & - & \textbf{97.86} & \textbf{94.96} & \underline{94.88} & \underline{78.58} 
& \underline{89.47} \\ 

\bottomrule
\end{tabular}%
\label{tabl:lga_impact_abl}
\vspace{-2mm}
\end{table}

\vspace{1mm}
\noindent\textbf{Vulnerable Points versus Vulnerable Patches.} 
To demonstrate the compatibility of vulnerable patches (VPatch) as compared to vulnerable points (Vpoint) with transformers, we report in Table~\ref{tabl:vpart_vpoint_abl} the obtained results when replacing VPatch with VPoint within LAA-Former and LAA-Swin. 
It can be observed that the use of VPatch not only results in better performance but also maintains a relatively lower computational cost as compared to VPoint. 
We note that the higher number of parameters and FLOPs associated with using VPoint is caused by a decoder designed to locate these points. Meanwhile, VPatch does not incur any decoders, making it more computationally efficient.

\vspace{1mm}
\noindent\textbf{Selection of $f_2$.} Table~\ref{tabl:vul_type_abl} compares two aggregation functions $f_2$ defined in Section~\ref{subsec:vulner_patch} coupled with LAA-Former: \textit{mean} and \textit{max} operations. In both cases, the stability of the results can be seen.  By default, we select the \textit{max} operation as it gives slightly better results.
In future works, we plan to investigate further selections of $f_2$, e.g., learnable alternatives.

\vspace{1mm}
\noindent\textbf{Learning Rate Sensitivity.}
In addition to the variations of patch size analyzed in Section~\ref{abl:patchsize_effects} and Section~\ref{sec:investigation}, we hypothesize that the learning rate (Lr) may also affect the learning capability of transformer architectures, especially when training with HQ pseudo-fakes such as SBI~\cite{sbi}, which enclose subtle forgeries. To analyze this, we keep the training protocol fixed and vary Lr values for LAA-Former, LAA-Swin, and their plain counterparts. The evaluation results on four datasets~\cite{ff++, celeb_df, dfd, dfdc} are reported in Table~\ref{tabl:lga_abl}. We observe that, for larger Lr values, the plain ViTs struggle to learn robust representations, leading to poor cross-dataset generalization. By contrast, the hierarchical design of Swin allows it to capture localized features more effectively and thus maintain relatively stable performance across all four datasets. Interestingly, integrating L2-Att consistently improves the generalizability of both ViT and Swin across all tested Lr values, with the gains being particularly noticeable at higher learning rates. This further highlights the impact of L2-Att in the context of deepfake detection.

\vspace{1mm}
\noindent\textbf{Impact of Loss Balancing Factor $\lambda_{att}$.} The weight $\lambda_{att}$ defined in Eq.~\eqref{equa:total_loss_transformer} is set empirically to $10$ as it yields the best performance on average. We report the results using different values of $\lambda_{att}$ within LAA-Former in Table~\ref{tabl:lga_impact_abl}. It can be observed that the generalization across four testing benchmarks remains robust regardless of the value of $\lambda_{att}$.

\vspace{-1mm}
\section{Conclusion}
\label{sec:conclu}

This paper introduces a unified, localized, artifact-aware attention learning framework called LAA-X for fine-grained deepfake detection. It aims at detecting HQ deepfakes while ensuring generalizability to unseen manipulations.
The main idea represents the introduction of a multi-task learning framework that incorporates auxiliary tasks, enforcing explicit attention to artifact-prone fine regions referred to as vulnerable regions. The latter are defined as the areas that are the most impacted by blending artifacts and are estimated by leveraging blending-based data synthesis techniques. We demonstrate that the proposed framework is architecture-agnostic and can be generalized to both CNN and Transformer architectures with small adaptations, resulting in two families, including LAA-Net and LAA-Former, respectively. Extensive evaluation and discussion on several challenging benchmarks demonstrate the superior performance of LAA-X as compared to SOTA methods. 
In future work, we will investigate strategies to extend the vulnerability concept to forgeries that do not necessarily exhibit blending artifacts, as well as to videos, to better capture spatio-temporal artifacts.

\vspace{3mm}
\noindent\textbf{\large{Acknowledgment}}
\label{sec:acknowledge}
\vspace{1mm}

This work is supported by the Luxembourg National Research Fund, under the BRIDGES2021/IS/16353350/FaKeDeTeR and UNFAKE, ref.16763798 projects, and by POST Luxembourg. Experiments were performed on the Luxembourg national supercomputer MeluXina. The authors gratefully acknowledge the LuxProvide teams for their expert support.


 

{
    \small
    \bibliographystyle{IEEEtran}
    \bibliography{main}
}


 
\begin{IEEEbiography}[{\includegraphics[width=1in,height=1.25in,clip,keepaspectratio]{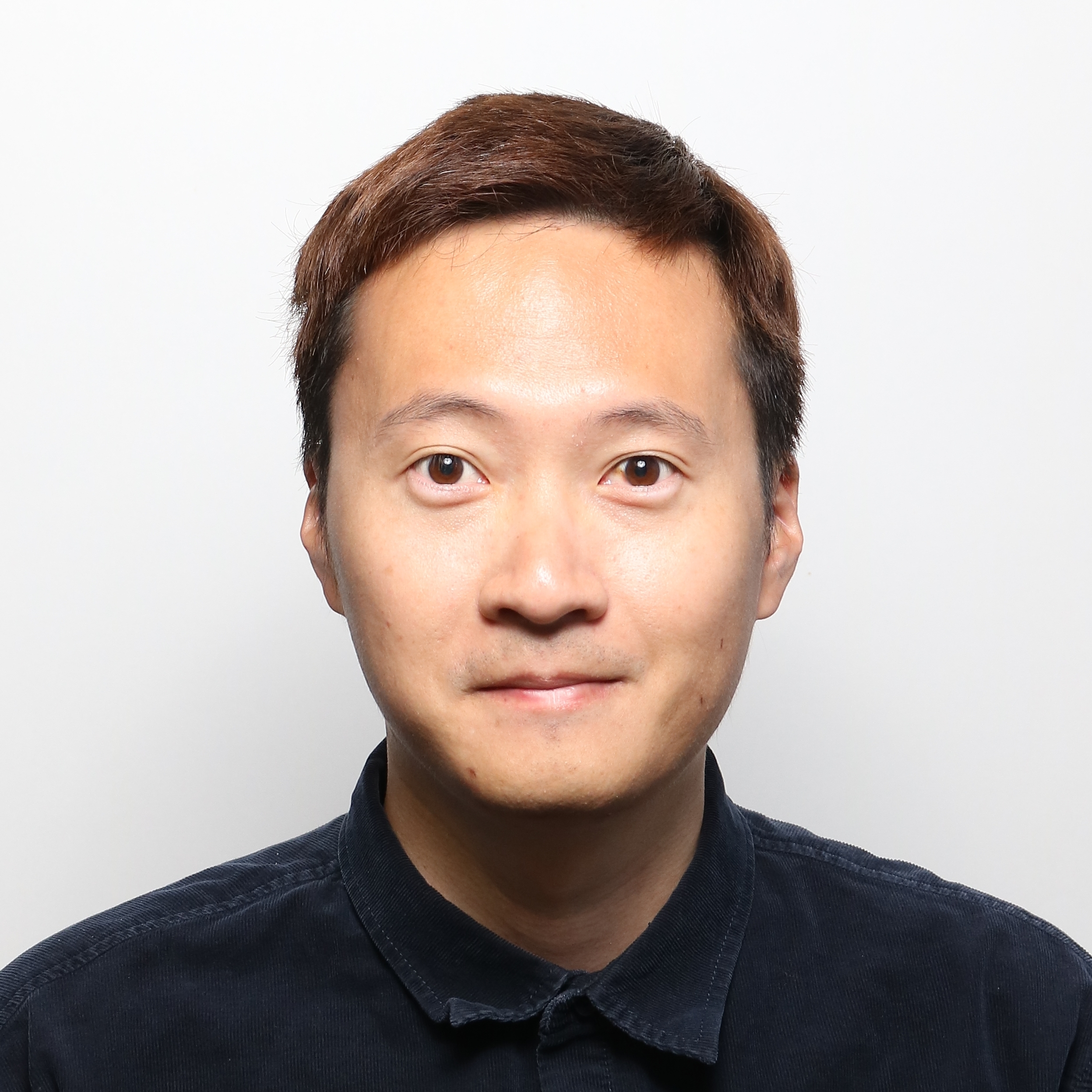}}]{Dat NGUYEN} received the Engineer's Degree in software engineering from the Military Technical Academy (MTA), Hanoi, Vietnam, and the MSc degree in computer science from the Vietnam National University (VNU), Hanoi, Vietnam. He is currently working toward the PhD degree at the Computer Vision, Imaging and Machine Intelligence (CVI2) Research Group, Interdisciplinary Centre for Security, Reliability and Trust (SnT), University of Luxembourg, Luxembourg. Prior to his PhD studies, he was a Senior Research Engineer at VinAI Research (now Qualcomm AI Research Vietnam), where he worked on AI perception models for autonomous driving. His research interests include computer vision, pattern recognition, and machine learning. He has authored papers at premier venues, including CVPR, ICCV, etc, and has served as a reviewer for these conferences.
\end{IEEEbiography}

\begin{IEEEbiography}[{\includegraphics[width=1in,height=1.25in,clip,keepaspectratio]{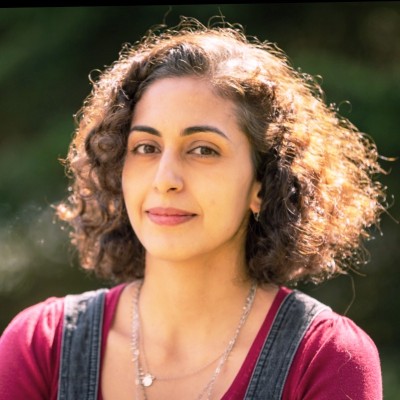}}]{Enjie GHORBEL} is an Assistant Professor at ENSI, University of Manouba, and a member of the CRISTAL laboratory. She is also a Research Fellow with the CVI2 research group at the Interdisciplinary Centre for Security, Reliability and Trust (SnT), University of Luxembourg. Prior to this role, she served as a Research Scientist at CVI2, SnT, University of Luxembourg until 2023. She obtained her HDR in 2025 and her PhD in Computer Science in 2017, both from the University of Rouen Normandie, as well as an engineering diploma from ENISO, University of Sousse, in 2014. Throughout her career, she has contributed to the acquisition and implementation of several national, international, and industrial research projects. Her research interests lie in computer vision and pattern recognition, with applications including human action recognition, deepfake detection, and pose estimation.
\end{IEEEbiography}

\begin{IEEEbiography}[{\includegraphics[width=1in,height=1.25in,clip,keepaspectratio]{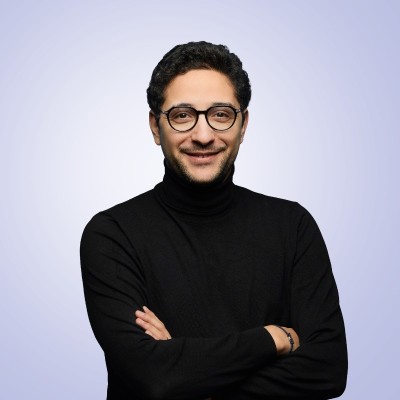}}]{Anis KACEM} is a permanent Research Scientist in Computer Vision, Imaging, and Machine Intelligence (CVI2) at the Interdisciplinary Centre for Security, Reliability and Trust (SnT), University of Luxembourg, and serves as Deputy Head of CVI2 research group. He holds a Computer Science Engineering degree from the National Institute of Applied Sciences and Technology (INSAT), Tunisia, obtained in 2014, and received his Ph.D. in Computer Science from the University of Lille, France, in 2018, with a thesis focused on geometric approaches for human behavior understanding from visual data. His research interests span computer vision and machine learning, with particular emphasis on 3D perception. He leads numerous research activities in close collaboration with industrial partners, fostering the transfer of advanced research into real-world applications. He has been a co-organizer of four editions of the SHARP workshop series, held in conjunction with ECCV 2020, CVPR 2021, CVPR 2022, and ICCV 2023, and has served on the technical program committees of several international workshops, including ManLearn (ICCV 2017), 3DHU (ICPR 2020), AI4Space (ECCV 2022 and CVPR 2024), and LFA (FG 2023). He regularly serves as a reviewer for top-tier AI and computer vision venues, including NeurIPS, CVPR, ECCV, and ICCV.
\end{IEEEbiography}

\begin{IEEEbiography}[{\includegraphics[width=1in,height=1.25in,clip,keepaspectratio]{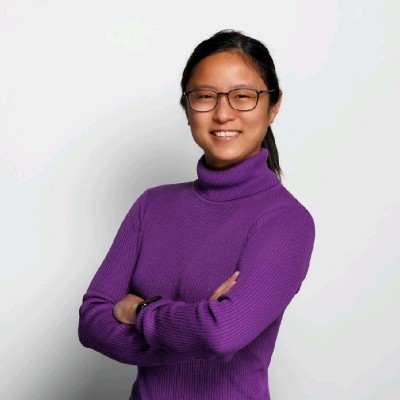}}]{Marcella ASTRID}
received her BEng in Computer Engineering from Multimedia Nusantara University, Tangerang, Indonesia, in 2015. She obtained her MEng in Computer Software and her PhD in Artificial Intelligence from the University of Science and Technology (UST), Daejeon, Korea, in 2017 and 2023, respectively. She was subsequently affiliated with the University of Luxembourg as a Postdoctoral Researcher from 2023 to 2025, where this research was conducted. She is currently affiliated with Helmholtz AI at Helmholtz Center Munich, Germany, as a health-focused AI consultant. Her current research interests include anomaly detection and computer vision.
\end{IEEEbiography}

\begin{IEEEbiography}[{\includegraphics[width=1in,height=1.25in,clip,keepaspectratio]{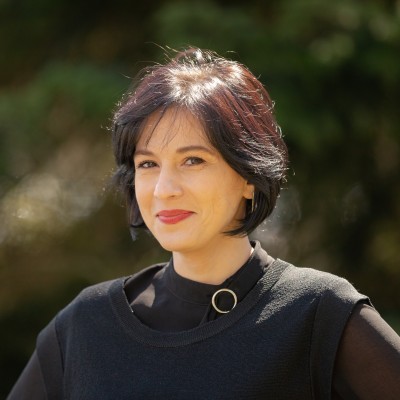}}]{Djamila AOUADA}
is Deputy Director at the University of Luxembourg’s Interdisciplinary Centre for Security, Reliability, and Trust (SnT). She is founder and head of the Computer Vision, Imaging and Machine Intelligence (CVI2) Research Group. She also heads the SnT Computer Vision Laboratory and co-heads the SnT Zero-G Laboratory. Having received her engineering degree in electronics from the École Nationale Polytechnique (ENP), Algiers, Algeria, and the Ph.D. degree in computer vision from North Carolina State University (NCSU), Raleigh, NC, USA, Djamila took up the challenge to come to Luxembourg to establish a leading research program in the field of computer vision at the newly founded SnT. Today her research group numbers some 30 researchers. She regularly serves the research community as a reviewer, Editor, Associate Editor, Chair, Area or Program Chair. She is the founder of the SHARP Workshop and chair of its four editions at ECCV 2020, CVPR 2021, CVPR 2022, and ICCV 2023. She has worked as a consultant for multiple renowned laboratories (Los Alamos National Laboratory, Alcatel Lucent Bell Labs., and Mitsubishi Electric Research Labs.). She co-authored over 150 scientific publications, has 4 patents and is the recipient of four IEEE best paper awards. She is a Senior member of the IEEE. Djamila is passionate about public outreach, particularly in promoting and encouraging women in STEAM. She served as the Chair of the IEEE Benelux Women in Engineering Affinity Group from 2014 to 2016, and has joined the Board of Directors of the Asteroid Foundation since April 2024. Since 2023, she has been appointed as member of the Algerian AI Board and the UK ART AI Board at the University of Bath for Accountable, Responsible and Transparent AI, further contributing to the advancement of AI initiatives. She served as awards chair in the last edition of EUVIP.
\end{IEEEbiography}


\clearpage
\newpage
\appendices
\section*{Supplementary Material}


\subsection{Self-Consistency Loss}

\begin{figure}[!h]
    \centering
    \includegraphics[width=\linewidth]{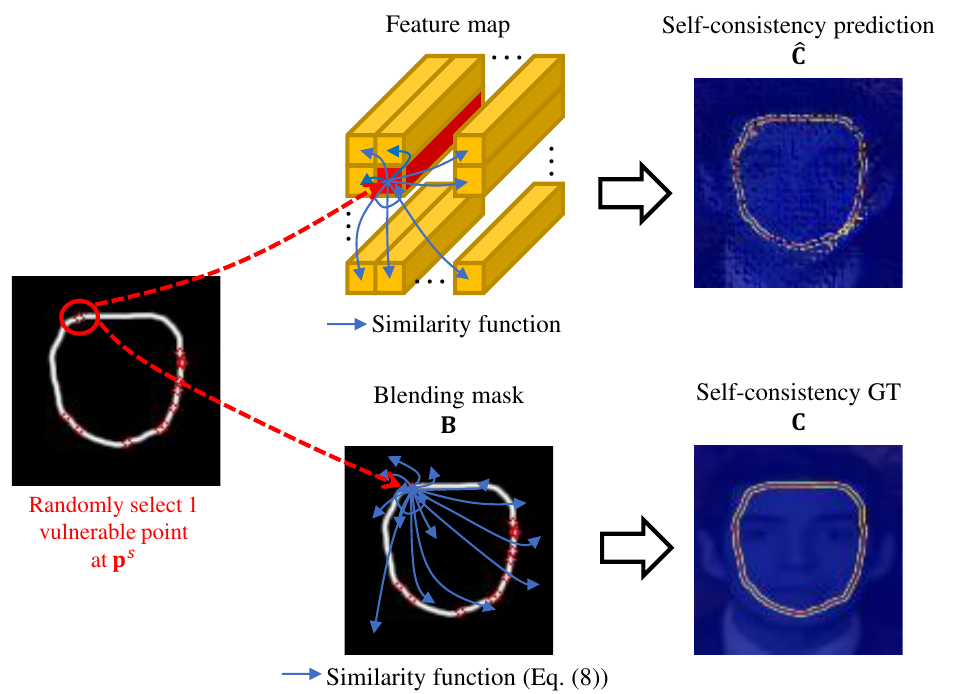}
    \caption{In order to generate the consistency map prediction $\hat{\mathbf{C}}$ as well as the associated ground truth $\mathbf{C}$, we first randomly select a vulnerable point located at $\mathbf{p}^s \in \mathcal P$. For computing $\hat{\mathbf{C}}$, we measure the similarity between the feature at $\mathbf{p}^s$ (red block) and the features generated from every point. Namely, we use the similarity function in \cite{cstency_learning}. As for $\mathbf{C}$, we measure the consistency values between the pixel at the $\mathbf{p}^s$ and all pixels in $\mathbf{B}$, as also described in Eq. (\textcolor{red}{7}) of the manuscript.}
    \label{fig:consistency}
\end{figure}

To clarify the calculation of the self-consistency loss, we show Figure \ref{fig:consistency}, which illustrates the generation process of the predicted and the ground-truth, $\hat{\mathbf{C}}$ and $\mathbf{C}$, respectively.  The self-consistency loss is a binary cross entropy loss between $\hat{\mathbf{C}}$ and $\mathbf{C}$.

\subsection{Ground Truth Generation of Heatmaps}

\begin{figure*}[!h]
    \centering
    \includegraphics[width=\linewidth]{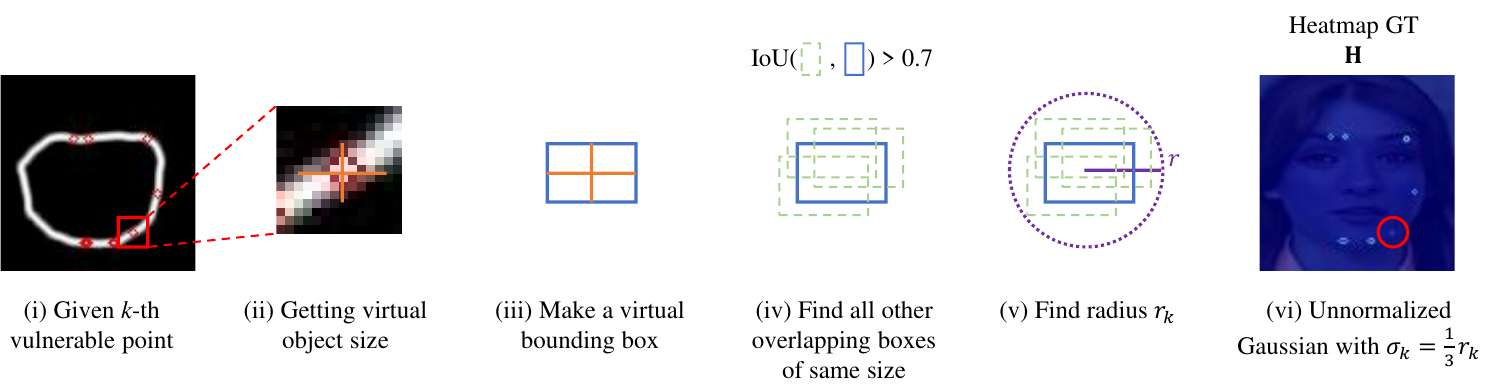}
    \caption{
    The generation process of ground truth heatmaps by producing using an \textit{Unnormalized Gaussian Distribution} given a selected vulnerable point.
    }
    \label{fig:heatmap_gt_generation}
\end{figure*}

In this section, we provide more details regarding the generation of ground-truth heatmaps, described in Section \textcolor{red}{III-B1-a} of the main paper. Firstly, a $k$-th vulnerable point, denoted as $\mathbf{p}^k$, is selected, as shown in Figure \ref{fig:heatmap_gt_generation} (i). Secondly, we measure the height and the width of the blending mask $\mathbf{B}$ at the point $\mathbf{p}^k$ shown as orange lines in Figure \ref{fig:heatmap_gt_generation} (ii). Using the calculated distances, a virtual bounding box is created, indicated by the blue box in Figure \ref{fig:heatmap_gt_generation} (iii). Then, we identify overlapping boxes, illustrated by dashed-line green boxes in Figure \ref{fig:heatmap_gt_generation} (iv), with the Intersection over Union (IoU) greater than a threshold ($t=0.7)$ compared to the virtual bounding box. A radius $r_k$ (solid purple line in Figure \ref{fig:heatmap_gt_generation} (v)) is calculated by forming a tight circle encompassing all these boxes. Finally, an \textit{Unnormalized Gaussian Distribution}, 
shown as a red circle in Figure \ref{fig:heatmap_gt_generation} (vi), is generated with a standard deviation $\sigma_k = \frac{1}{3}r_k$ (Eq. (\textcolor{red}{4}) of the manuscript).
The steps are repeated for every vulnerable point $k \in [\![ 1, \text{card}(\mathcal P)]\!]$. The final $\mathbf{H}$ is the superimposition of all  $g_{ij}^k$.

\begin{table*}
\centering
\caption{\textbf{In-dataset and Cross-dataset evaluation} in terms of AUC (\%), AP (\%), AR (\%), and mF1 (\%) at the \textit{video-level} on multiple deepfake datasets. Results for comparison are directly extracted from the original papers. $\ast$ indicates our reproduced results using official pre-trained weights. \textbf{Bold} and \underline{Underlined} highlight the best and the second-best performance, respectively.}
\resizebox{\linewidth}{!}{
{\rowcolors{30}{Plum!10}{SkyBlue!20}
\begin{tabular}{c c cc c HHHH cccc cccc cccc cccc cccc}
\toprule
\multirow{3}{*}{Method} & \multirow{3}{*}{Venue} & \multicolumn{2}{c}{Training set} & \multicolumn{25}{c}{Test set} \\
\cmidrule(lr){3-4}
\cmidrule(lr){5-29}
& & \multirow{2}{*}{Real} & \multirow{2}{*}{Fake} & FF++ & \multicolumn{4}{H}{CDF1} & \multicolumn{4}{c}{CDF2} & \multicolumn{4}{c}{DFW} & \multicolumn{4}{c}{DFD} & \multicolumn{4}{c}{DFDCP} & \multicolumn{4}{c}{DFDC} \\
\cmidrule(lr){5-6}\cmidrule(lr){10-13}\cmidrule(lr){14-17}\cmidrule(lr){18-21}\cmidrule(lr){22-25}\cmidrule(lr){26-29}
 & & & & AUC & AUC & AP & AR  & mF1  & AUC & AP & AR & mF1 & AUC & AP & AR & mF1 & AUC & AP & AR & mF1 & AUC & AP & AR & mF1 & AUC & AP & AR & mF1 \\
\midrule
\midrule
Xception$^\ast$~\cite{ff++} & ICCV'19 & $\checkmark$ & $\checkmark$ & 93.60 & 58.81 & 65.59 & 55.58 & {60.17} & 61.18 & 66.93 & 52.40 & {58.78} & 65.29 & 55.37 & 57.99 & {56.65} & 89.75 & 85.48 & 79.34 & 82.29 & 69.90 & 91.98 & 67.07 & 77.57 & 58.98 & 55.32 & 55.84 & 55.58 \\

FaceXRay+BI~\cite{fxray} & CVPR'20 & $\checkmark$ & $\checkmark$ & 99.20 & 80.58 & 73.33 & - & - & 79.50 & - & - & {-} & - & - & - & {-} & 95.40 & 93.34 & - & - & 65.50 & - & - & - & - & - & - & - \\




Multi-attentional$^\ast$~\cite{multi-attentional} & CVPR'21 & $\checkmark$ & $\checkmark$ & 95.32 & 69.14 & 74.03 & 52.70 & {61.57} & 68.26 & 75.25 & 52.40 & {61.78} & 73.56 & 73.79 & 63.38 & {68.19} & 92.95 & 96.51 & 60.76 & 74.57 & 83.81 & 96.52 & \underline{77.68} & \underline{86.08} & 70.05 & 67.11 & 63.53 & 65.27 \\

PCL+I2G~\cite{cstency_learning} & ICCV'21 & $\checkmark$ & $\times$ & 99.11 & \textbf{98.30} & - & - & - & 90.03 & - & - & - & - & - & - & - & 99.07 & - & - & - & 74.27 & - & - & - & 67.52 & - & - & - \\

RECCE$^\ast$~\cite{ete_recons} & CVPR'22 & $\checkmark$ & $\checkmark$ & 99.56 & 49.96 & 63.04 & 50.87 & {56.31} & 70.93 & 70.35 & 59.48 & 64.46 & 68.16 & 54.41 & 56.59 & {55.48} & 98.26 & 79.42 & 69.57 & 74.17 & 80.98 & 92.75 & 70.69 & 80.23 & 71.19 & 68.97 & 63.53 & 66.14 \\


SBI$^\ast$~\cite{sbi} & CVPR'22 & $\checkmark$ & $\times$ & 98.23 & 82.65 & 77.09 & 70.68 & 73.75 & 85.55 & 77.81 & 68.13 & 72.65 & 67.47 & 55.87 & 55.82 & 55.85 & 96.04 & 92.79 & \underline{89.49} & 91.11 & 82.22 & 93.24 & 71.58 & 80.99 & 69.77 & 72.25 & 54.87 & 62.37 \\

DFDT~\cite{dfdt} & Appl.Sci.'22 & $\checkmark$ & $\checkmark$ & 97.9 & - & - & - & - & 88.3 & - & - & - & - & - & - & - & - & - & - & - & 76.1 & - & - & - & - & - & - & - \\

SFDG~\cite{sfdg} & CVPR'23 & $\checkmark$ & $\checkmark$ & 99.53 & - & - & - & {-} & 75.83 & - & - & {-} & 69.27 & - & - & {-} & 88.00 & - & - & - & 73.63 & - & - & - & - & - & - & - \\



CADDM$^\ast$~\cite{caddm} & CVPR'23 & $\checkmark$ & $\checkmark$ & 99.26 & 89.36 & 93.25 & 81.41 & 86.93 & 80.70 & 87.72 & 72.56 & 79.42 & 76.31 & 79.19 & \underline{69.35} & \underline{73.95} & 99.03 & \underline{99.59} & 82.17 & 90.04 & 71.00 & 95.60 & 68.49 & 79.81 & 70.33 & 70.01 & \underline{63.60} & 66.65 \\

AUNet~\cite{aunet} & CVPR'23 & $\checkmark$ & $\times$ & 99.46 & - & - & - & {-} & 92.77 & - & - & {-} & - & - & - & {-} & \underline{99.22} & - & - & - & 86.16 & - & - & - & 73.82 & - & - & - \\



LSDA~\cite{LSDA} & CVPR'24 & $\checkmark$ & $\checkmark$ & 95.8 & - & - & - & - & 89.8 & - & - & - & 75.6 & - & - & - & 95.6 & - & - & - & 81.2 & - & - & - & 73.5 & - & - & - \\

FA-ViT~\cite{FAViT} & TCSVT'24 & $\checkmark$ & $\checkmark$ & 99.6 & - & - & - & - & 93.83 & - & - & - & \bf{84.32} & - & - & - & 94.88 & - & - & - & 85.41 & - & - & - & \underline{78.32} & - & - & - \\

UDD~\cite{UDD} & AAAI'25 & $\checkmark$ & $\checkmark$ & - & - & - & - & - & 93.1 & - & - & - & - & - & - & - & 95.5 & - & - & - & 88.1 & - & - & - & - & - & - & - \\


FreqDebias~\cite{debias_deepfake} & CVPR'25 & $\checkmark$ & $\checkmark$ & - & - & - & - & - & 89.6 & - & - & - & - & - & - & - & - & - & - & - & - & - & - & - & 77.8 & - & - & - \\

\midrule

AltFreezing~\cite{altfreezing} & CVPR'23 & $\checkmark$ & $\checkmark$ & 98.60 & - & - & - & - & 89.50 & - & - & {-} & - & - & - & - & 98.50 & - & - & - & 70.84 & - & - & - & 71.74 & - & - & - \\

ISTVT~\cite{istvt} & TIFS'23 & $\checkmark$ & $\checkmark$ & 99.0 & - & - & - & - & 84.1 & - & - & - & - & - & - & - & - & - & - & - & 74.2 & - & - & - & - & - & - & - \\

TALL-Swin~\cite{tall_swin} & ICCV'23 & $\checkmark$ & $\checkmark$ & 99.87 & - & - & - & - & 90.79 & - & - & - & - & - & - & - & - & - & - & - & 76.78 & - & - & - & - & - & - & - \\

FakeSTormer~\cite{fakestormer} & ICCV'25 & $\checkmark$ & $\times$ & 98.4 & - & - & - & - & 92.4 & - & - & - & 74.2 & - & - & - & 98.5 & - & - & - & 90.0 & - & - & - & 74.6 & - & - & - \\

\midrule
\midrule
LAA-Net (Ours w/ BI) & CVPR'24 & $\checkmark$ & $\times$ & \underline{99.95} & 92.46 & 95.54 & 50.01 & 65.64 & 86.28 & 91.93 & 50.01 & 64.78 & 57.13 & 56.89 & 50.12 & {{53.29}} & \textbf{99.51} & \textbf{99.80} & \textbf{95.47} & \textbf{97.59} & 69.69 & 93.67 & 50.12 & 65.30 & 71.36 & 73.02 & 55.82 & 63.27 \\ 

LAA-Former (Ours w/ BI) & - & $\checkmark$ & $\times$ & 99.23 & & & & & 90.34 & 94.90 & 63.38 & 76.00 & 72.62 & 75.98 & 59.97 & 67.03 & 93.42 & 97.49 & 77.26 & 86.21 & 78.71 & 96.23 & 60.17 & 74.04 & 76.84 & \textbf{80.82} & 63.31 & \underline{71.00} \\

\midrule 
LAA-Net (Ours w/ SBI) & CVPR'24 & $\checkmark$ & $\times$ & \textbf{99.96} & \underline{93.11} & \textbf{95.64} & \textbf{89.78} & {\textbf{92.62}} & \textbf{95.40} & \textbf{97.64} & \textbf{87.71} & {\textbf{92.41}} & 80.03 & \underline{81.08} & 65.66 & 72.56 & 98.43 & 99.40 & 88.55 & \underline{93.64} & \underline{86.94} & \underline{97.70} & 73.37 & 83.81 & 72.43 & 74.46 & 57.39 & 64.81 \\

LAA-Former (Ours w/ SBI) & - & $\checkmark$ & $\times$ & 97.67 & & & & & \underline{94.45} & \underline{97.15} & \underline{81.29} &  \underline{88.51} & \underline{81.74} & \textbf{83.72} & \bf{71.44} & \bf{77.10} & 96.12 & 98.31 & 78.85 & 87.52 & \textbf{96.30} & \textbf{99.50} & \bf{78.01} & \bf{87.45} & \textbf{78.91} & \underline{80.01} & \bf{70.86} & \bf{75.15} \\

\bottomrule
\end{tabular}%
}}
\label{tabl:cross_auc_full_metrics_supp}
\end{table*}

\subsection{Additional Results}

In addition to AUC, we provide results using additional metrics, namely, Average Precision (AP), Average Recall (AR), Accuracy (ACC), and mean F1-score (mF1). 

Table~\ref{tabl:ff_full_metrics_suppl1} and Table~\ref{tabl:cross_auc_full_metrics_supp} report the results under the in-dataset and the cross-dataset settings, respectively. Overall, it can be seen that LAA-Net and LAA-Former achieve better performance than other state-of-the-art methods.

\begin{table}[]
\centering
\caption{In-dataset evaluation on FF++~\cite{ff++} reported by ACC, AUC, AP, AR, and mF1.}
\label{tabl:ff_full_metrics_suppl1}
\resizebox{\columnwidth}{!}{%
\begin{tabular}{c|cc|cccccc}
\hline
\multirow{2}{*}{Method} & \multicolumn{2}{c|}{Training Set} & \multicolumn{5}{c}{FF++ Test Set~\cite{ff++}} \\
\cline{2-8}
 & Real & Fake & ACC & AUC & AP & AR & mF1 \\ 
\hline
\hline
Ours w/ BI~\cite{fxray} & \checkmark & $\times$ & 99.03 & 99.95 & 99.99 & 99.21 & 99.60 \\
Ours w/ SBI~\cite{sbi} & \checkmark & $\times$ & 99.04 & 99.96 & 99.99 & 99.29 & 99.64 \\
\hline
\end{tabular}%
}
\end{table}

\vspace{1mm}
\subsubsection{Qualitative Results - Gaussian Noise}

\begin{figure}
    \centering
    \includegraphics[width=\linewidth]{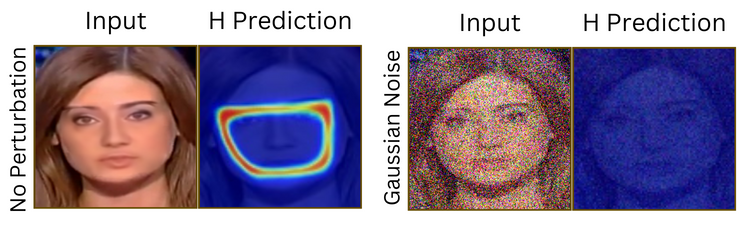}
    \caption{Detection of vulnerable points w/o and w/ Gaussian noise.}
    \label{fig:noise-refer}
\end{figure}

In Table~\textcolor{red}{IV} of the main manuscript, the performance of LAA-Net declined significantly when encountering Gaussian Noise perturbations. One possible reason is that the introduction of noise elevates the difficulty of detecting the vulnerable points. To confirm that, we report the inference of the heatmap before and after applying a Gaussian Noise on a facial image in Figure~\ref{fig:noise-refer}. As it can be observed, the detection of vulnerable points is highly impacted by the introduction of a Gaussian noise.

\vspace{1mm}
\subsubsection{Robustness to Compression}

To assess the robustness of LAA-Net to compression, we test LAA-Net on the c23 version of FF++, and the overall AUC is equal to $89.30$\%.


\subsection{More Details regarding the training Setup in Section~\textcolor{red}{III-C1}}

In this section, we present more details related to the experimental settings in Section~\textcolor{red}{III-C1} of the main paper. 


In Figure~\textcolor{red}{6}-b, all models, including CNNs and variants of ViTs are trained on FF++~\cite{ff++} with both real and fake data for $50$ epochs. Following conventional spitting~\cite{ff++}, we uniformly extract $128$ and $32$ frames of each video for training and validation, respectively. Hence, there are in total of $460800$ and $22400$ frames for the corresponding task. The weights of models are initialized by pretrained on ImageNet~\cite{imagenet}. We employ different optimizers as Adam is often used with CNNs~\cite{sladd, ost, caddm, fxray, multi-attentional, sfdg} and AdamW with ViTs~\cite{swin, twin, deit, PVT}. The learning rate is initially set to $10^{-4}$ and linearly decays to $0$ at the end of the training period. All experiments are carried out using a NVIDIA A100 GPU.




\subsection{Architecture Details}
We describe in detail the hyperparameters of the two considered LAA-Former variants as follows:
\begin{itemize}
    \item LAA-Former-S: $H=W=112$, $P=8$, $L=12$, $D=384$, MLP size $=1536$, No. Heads $=6$, Params$=23$M, FLOPs$=8.9$G.
    \item LAA-Former-B: $H= W=224$, $P=16$, $L=12$, $D=768$, MLP size $=3072$, No. Heads $=12$, Params$=91$M, FLOPs$=35.8$G.
\end{itemize}
where the \textit{MLP size} represents the dimension of hidden layers in MLP, the \textit{No. Heads} denotes the number of heads in MHSA, \textit{Params} is the number of parameters, and \textit{FLOPs} represents the computational cost in terms of floating point operations per second. 

For LAA-Swin architecture, we adopt these two backbone variants from Swin~\cite{swin}, namely:
\begin{itemize}
    \item LAA-Swin-S: $H=W=224$, $P=4$, $M=7$, $d=32$, $\alpha=4$, $C_h=96$, Layer Numbers = \{2, 2, 18, 2\}, No. Heads = \{3, 6, 12, 24\}, Params$=55$M, FLOPs$=6.5$G.
    \item LAA-Swin-B: $H=W=224$, $P=4$, $M=7$, $d=32$, $\alpha=4$, $C_h=128$, Layer Numbers = \{2, 2, 18, 2\}, No. Heads = \{4, 8, 16, 32\}, Params$=91$M, FLOPs$=11.5$G.
\end{itemize}
where $M$ is the window size, $d$ is the query dimension of each head, the expansion layer of each MLP is $\alpha$, and $C_h$ denotes the channel number in the hidden layers during the first stage.

\subsection{More Details regarding the Datasets}
\label{subsec:dataset_supp}

\noindent\textbf{Datasets.} 
The FF++~\cite{ff++} dataset is used for training and validation. In our experiments, we follow the standard splitting protocol of~\cite{ff++}. This dataset contains $1000$ original videos and $4000$ fake videos generated by four different manipulation methods, namely, Deepfakes (DF)~\cite{deepfake}, Face2Face (F2F)~\cite{face2face}, FaceSwap (FS)~\cite{faceswap}, and NeuralTextures (NT)~\cite{neutex}. In the training process, we utilize only real images to dynamically generate pseudo-fakes, as discussed in Section~\textcolor{red}{III} of the main paper. To evaluate the generalization capability of the proposed approach as well as its robustness to high-quality deepfakes, we follow the cross-dataset setting on seven challenging datasets encompassing deepfakes of varying quality and diverse manipulation techniques: (1) \textbf{Celeb-DFv2}~\cite{celeb_df} (CDF2), a well-known benchmark with high-quality deepfakes; (2) \textbf{Google DeepFake Detection}~\cite{dfd} (DFD), which includes $3,000$ forged videos featuring $28$ actors in various scenes; (3) \textbf{DeepFake Detection Challenge}~\cite{dfdc} (DFDC) and its preview version (i.e., (4) \textbf{DeepFake Detection Challenge Preview}~\cite{dfdcp} (DFDCP)), a large-scale dataset containing numerous distorted videos with issues such as compression and noise; (5) \textbf{WildDeepfake}~\cite{wdf} (DFW), a dataset fully sourced from the internet without prior knowledge of manipulation methods; (6) a diffusion-based test set \textbf{DiffSwap}~\cite{diffusionface}, and (7) \textbf{DF40}~\cite{df40}, a highly diverse and large-scale dataset comprising $40$ distinct deepfake techniques, enables more comprehensive evaluations for the next generation of deepfake detection.

To assess the quality of the considered datasets, we compute the Mask-SSIM~\cite{mssim_pose} for each benchmark. In particular, CDF2~\cite{celeb_df} is formed by the most realistic deepfakes with an average Mask-SSIM value of $0.92$, followed by DFD, DF40, DFDC, and DFDCP with an average Mask-SSIM of $0.88$, $0.87$, $0.84$ and $0.84$, respectively. We note that computing the Mask-SSIM~\cite{celeb_df} for DFW and DiffSwap was not possible since real and fake images are not paired.

\vfill

\end{document}